\documentclass[lettersize,journal]{IEEEtran}
\usepackage{amsmath,amsfonts}
\usepackage{algorithmic}
\usepackage{algorithm}
\usepackage{array}
\usepackage{textcomp}
\usepackage{stfloats}
\usepackage{url}
\usepackage{verbatim}
\usepackage{graphicx}

\usepackage{booktabs}

\usepackage[commandnameprefix=always]{changes}
\newcommand{\review}[1]{{\leavevmode\color{black}#1}}
\usepackage{todonotes}

\usepackage{multirow}
\usepackage{makecell}
\usepackage{caption}
\usepackage{subcaption}

\def\eg{\emph{e.g}. } \def\Eg{\emph{E.g}. }
\def\ie{\emph{i.e}. } 
\def\cf{\emph{c.f}. }

 \def\etal{\emph{et al}. }

\newcommand{\rev}[1]{\textcolor{black}{#1}}

\newcolumntype{L}[1]{>{\centering\arraybackslash}m{#1}}

\begin{document}

\title{MCLFIQ:\\ Mobile Contactless Fingerprint Image Quality}

\author{\IEEEauthorblockN{Jannis Priesnitz$^1$, Axel Weißenfeld$^2$, Laurenz Ruzicka$^2$, Christian Rathgeb$^1$, Bernhard Strobl$^2$, Ralph Lessmann$^3$, Christoph Busch$^1$} \\
\IEEEauthorblockA{1 -- da/sec – Biometrics and Security Research Group, Hochschule Darmstadt, Germany\\}
\IEEEauthorblockA{2 -- AIT Austrian Institute of Technology, Vienna, Austria\\}
\IEEEauthorblockA{}{3 -- HID Global, Jena, Germany  \\}
}



\maketitle

\begin{abstract}
We propose \textit{MCLFIQ: Mobile Contactless Fingerprint Image Quality}, the first quality assessment algorithm for mobile contactless fingerprint samples. To this end, we re-trained the NIST Fingerprint Image Quality (NFIQ) 2 method, which was originally designed for contact-based fingerprints, with a synthetic contactless fingerprint database. We evaluate the predictive performance of the resulting  MCLFIQ model in terms of Error-vs.-Discard Characteristic (EDC) curves on three real-world contactless fingerprint databases using three recognition algorithms. In experiments, the MCLFIQ method is compared against the original NFIQ 2.2 method, a sharpness-based quality assessment algorithm developed for contactless fingerprint images \rev{and the general purpose image quality assessment method  BRISQUE. Furthermore, benchmarks on four contact-based fingerprint datasets are also conducted.}

Obtained results show that the fine-tuning of NFIQ 2 on synthetic contactless fingerprints is a viable alternative to training on real databases. Moreover, the evaluation shows that our MCLFIQ method works more accurate and robust compared \rev{ to all baseline methods on contactless fingerprints.} We suggest considering the proposed MCLFIQ method as a \rev{starting point for the development of} a new standard algorithm for contactless fingerprint quality assessment. 
\end{abstract}


\section{Introduction}
\label{sec:intor}
In the past years, contactless fingerprint recognition has been introduced as a more convenient alternative to contact-based schemes \cite{priesnitz2021overview, jcp2030036, 9573458}. In contrast to contact-based capturing schemes where the finger is pressed onto a planar surface, contactless recognition workflows do not require any contact between the subject and the capturing subsystem. This avoids distinct problems like low contrast caused by dirt, humidity on the capturing device, or latent fingerprints. Moreover, contactless fingerprint recognition schemes typically have a higher user acceptance, especially in multi-user scenarios, where different individuals share one capture device. In said cases, the subjects might have fewer hygienic concerns using contactless fingerprint recognition \cite{priesnitz2021mobile}. 

A wide variety of different capturing setups have been developed for contactless fingerprint recognition. The range of capturing devices reaches from expensive stationary devices capturing 3D samples to lightweight mobile setups. However, it can be observed from the literature that most contactless fingerprint capturing devices are mobile handheld devices like smartphones \cite{priesnitz2021mobile, priesnitz2021overview, 9573458}.
The contactless fingerprint recognition workflow, especially in mobile capturing scenarios, suffers from an inferior biometric performance, which is mainly caused by a more challenging capturing process. External influences like illumination or the distance between the capturing device and the fingertip have a major impact on the quality of a captured sample.


\begin{figure}[!t]
	\centering
    \begin{subfigure}[b]{0.225\textwidth}
    \includegraphics[width=0.45\linewidth]{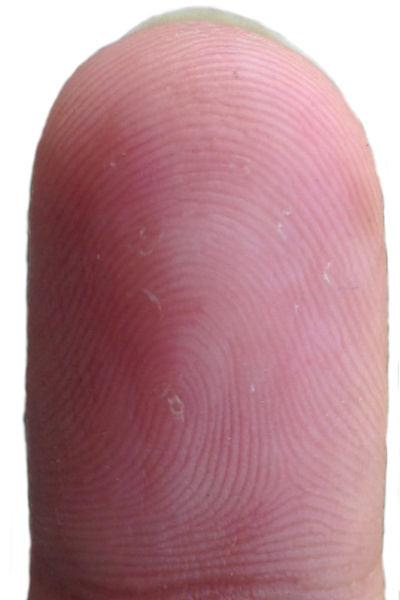}
    \hfil
    \includegraphics[width=0.45\linewidth]{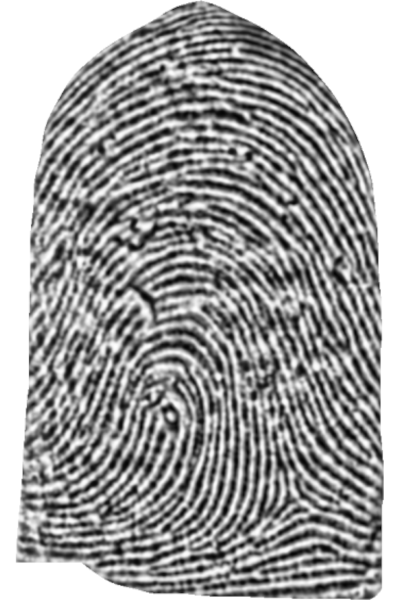}
    \caption{MCLFIQ score: 78}
    \label{fig:example_real_high_o}
    \end{subfigure}
    \hfil
    \begin{subfigure}[b]{0.225\textwidth}
    \includegraphics[width=0.45\linewidth]{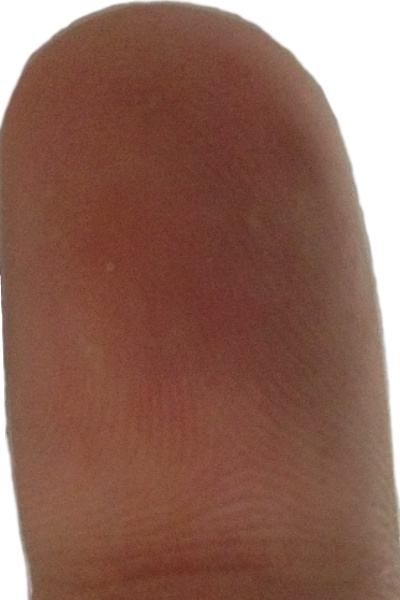}
    \hfil
    \includegraphics[width=0.45\linewidth]{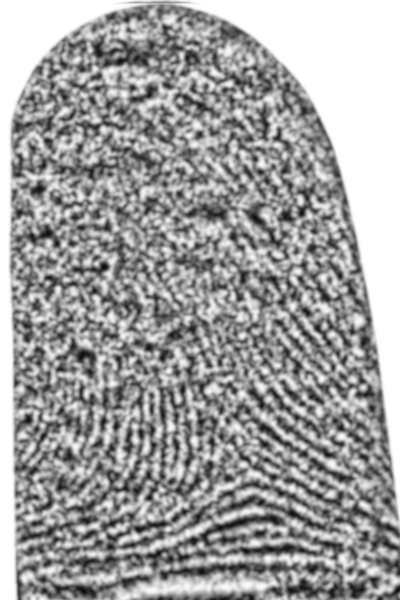}
    \caption{MCLFIQ score: 4}
    \label{fig:example_real_low_o}
    \end{subfigure}
	\caption{Two fingerprint samples with pre-processed images of same subject: (a) high quality, (b) low quality.}
	\label{fig:example_real}
\end{figure}

\rev
The signals obtained from a mobile contactless fingerprint capturing device are most commonly 2D color images, which cannot be directly fed into a recognition workflow. Firstly, an elaborated pre-processing is required in order to transfer the captured finger image to a contactless fingerprint sample. This task typically includes steps like gray scale conversion, ridge-line enhancement, normalization and rotation \cite{priesnitz2021overview}. 

Both, the capturing subsystem and the pre-processing can negatively impact the recognition performance. Figure \ref{fig:example_real} illustrates two contactless fingerprints of the same subject obtained from a smartphone-based capturing setup. Figures \ref{fig:example_real_high_o}, \ref{fig:example_real_low_o} depict the segmented finger image together with the corresponding final contactless fingerprint sample after enhancement. 
From Figure \ref{fig:example_real} we can see that the two samples are of different quality:  Figure \ref{fig:example_real_high_o} of rather high quality and Figure \ref{fig:example_real_low_o} of low quality.
The low-quality sample could be the result of a capturing attempt in a challenging environmental situation. The resulting finger image lacks a visible ridge-line characteristic and is not usable for further processing.
For this reason, a precise and robust quality assessment tool for contactless fingerprints is of high interest to assess if a candidate sample is of sufficient quality for a recognition workflow. Further, actionable feedback may be provided to the capture subject or the biometric attendant to initiate a re-capture of the finger image.

Quality in general is defined as "being suitable for the intended purpose" or the "fitness for purpose". With a proper quality assessment, system operators want to ensure that their service operates as specified. To achieve a reliable and reproducible attribution of quality, operational definitions should be established to achieve an objective and automated quality assessment. 
Within the context of biometric recognition, quality assessment refers to the mapping of an individual biometric signal to a numerical value, whereas higher values indicate a better quality and thus predict a stronger recognition performance. 
The ISO/IEC 29794-1 \cite{ISO-IEC-29794-1-QualityFramework-160915} defines guidelines for a utility-based biometric sample quality assessment, which includes three aspects:

\begin{itemize}
\item \textbf{Character:} Character is an expression of quality based on the inherent properties of the source from which the biometric sample is derived. \Eg a scarred finger has a poor character.
\item \textbf{Fidelity:} Fidelity reflects the degree of the sample similarity to its source. \Eg a capturing device with low resolution captures a sample of low fidelity.
\item \textbf{Utility:} The utility indicates the predicted influences of a sample to the recognition performance of a biometric system. Biometric utility depends on the character and the fidelity of a sample.
\end{itemize}
\label{items:utilityetc}
\rev{It should be noted that a clear separation of the influence between character and fidelity is not possible in many cases. \Eg a dirty finger is neither attributed to character nor fidelity. Furthermore, destinct identification workflows might be more robust against challenges regarding either character or fidelity than others. For this reason, ISO/IEC 29794-1 \cite{ISO-IEC-29794-1-QualityFramework-160915} suggests considering utility for sample quality assessment.    }

Figure \ref{fig:example_real} illustrates both, the impact of character and fidelity on a contactless fingerprint sample. In Figure \ref{fig:example_real_high_o} some artifacts of death flaking off skin can be seen which refer to the character. However, in this case, the impact of character on the utility is minor.
The sample depicted in \ref{fig:example_real_low_o} has a rather low contrast, which impacts the fidelity. From the corresponding pre-processed image, it is observable that barely  any ridge line characteristic is extractable.
Sample quality assessment is of major importance for increasing the performance of a biometric system. A quality assessment algorithm ensures that only samples of high quality are processed in the biometric system. An accurate and robust quality assessment algorithm usually enables a biometric system to be operated at lower error rates. This will in turn enhance the system's security associated with the False Acceptance Rate (FAR) and user comfort associated with the False Rejection Rate (FRR). 

For contact-based fingerprint recognition, NFIQ 2 is the reference implementation to ISO/IEC 29794-4 which has been established as the de facto standard \cite{bausinger2011fingerprint}. 
%
NFIQ 2 is an open-source software which was implemented under the leadership of the National Institute of Standards and Technology (NIST). The software links image quality of optical and ink scanned 500 PPI fingerprints to operational recognition performance. NFIQ 2 consists of 74 quality features which are formally standardized in ISO/IEC 29794-4 \cite{ISO29794-4}. A random forest classifier maps the individual quality measures to a unified quality score in the range [0,100].  

\rev{
It should be noted that \textit{NFIQ~2} refers to the general redesign of NFIQ as documented in NISTIR 8382 \cite{Tabassi-NFIQ21-NISTIR-8382-2021}, whereas \textit{NFIQ~2.0 -- NFIQ~2.2} refer to a specific release\footnote{\url{https://github.com/usnistgov/NFIQ2}}. 
}

In 2021, a virtual Workshop on Fingerprint Image Quality (NFIQ 2.1) was organized by the European Association for Biometrics (EAB) in cooperation with NIST and other institutions \footnote{https://eab.org/events/program/248}. Throughout this workshop, the importance of a reliable quality assessment for fingerprint images was emphasized. Moreover, the speakers and panelists formulated the interest of extending the scope of NFIQ 2 to other capturing technologies like contactless fingerprints. Up to now, no proposal of such an algorithm has yet been made. The main reasons are the lack of a suitable database for training machine learning algorithms. 

\begin{table*}[t]
\caption{Overview on published works on the topic of contactless fingerprint quality assessment. It should be noted that no information is available on the internal functionality of the Verifinger quality metric used by Wild \etal.}
\centering
\scriptsize
\begin{tabular}{cccccc}
	\hline
	\textbf{Author} & \textbf{Year} & \textbf{Method} & \textbf{Features} & \textbf{Classifier} & \textbf{Database} \\
	\hline
	Parziale and Chen \cite{Parziale2009} & 2009 & Block-based quality features (non-binary) & \makecell{Block based coheriance measure \\ based on local gradience} & none & own \\
	Labati \etal \cite{labati2010neural} & 2010 & Assessment of positional inaccuracies & \makecell{Distance between finger and \\ capturing device, rotation, movement} & Neural net & own \\
	Li \etal \cite{li2013quality} & 2013 & \makecell{Quality features on \\ unsegmented images} & \makecell{45 features representing different \\ aspects like finger-to-capturing \\ device distance, rotation, movement} & SVM & own\\
	Li \etal \cite{li2013autocorrelation} & 2013 & \makecell{Block-based quality assessment with \\ binary decision for each block} & \makecell{6 quality features incl. ridge \\ spatial frequency characteristics, certainty\\ of orientation and minutiae count} & SVM & own \\
	Liu \etal \cite{7566671} & 2016 & \makecell{Extraction of general image-based attributes \\ for contactless fingerprint, face and iris} & \makecell{Contrast, sharpness, luminance, \\artifacts and color (generic assessment)} & none & none\\
	Wild \etal \cite{wild2019comparative} & 2019 & \makecell{Evaluation of NFIQ 1.0, NFIQ 2.0, \\ Verfinger} & Features included in NFIQ 1, NFIQ 2 & \makecell{SVM, \\Random forest} & own \\
	Priesnitz \etal \cite{9211015}& 2020 & \makecell{Evaluation of NFIQ 2.0 on \\ contactless fingerprints} & 74 features included in NFIQ 2 & Random forest & \makecell{ISPFDv1, CL2CB \\ PolyU, FVC06, MCYT} \\
	Kauba \etal \cite{kauba2021towards} & 2021 & \review{Global image s}harpness & \makecell{Canny-filter-based  edge detection} & none & \makecell{own (as part of \\ database capturing)} \\
	\hline
\end{tabular}

\label{tab:relatedwork}

\end{table*}

\subsection{Contribution}
In this work, we address the aforementioned demands and present \textit{MCLFIQ}, the first quality assessment algorithm for contactless fingerprint images captured with mobile devices. 
MCLFIQ represents an adaptation of the NFIQ 2 framework for contactless fingerprint images:

\begin{itemize}
\item First, we define framework conditions for pre-processing contactless fingerprint images in order to achieve sample consistency and to make all images suitable for quality assessment using our method.
\item We then iteratively re-trained the random forest included in the NFIQ 2 framework using a synthetically generated contactless fingerprint database. Here, with every iteration step, the amount and appearance of the training data is adjusted in order to optimize the evaluation results.
\item Finally, we test our newly generated model, referred to as MCLFIQ. For this, we use three real world contactless fingerprint databases, the ISPFDv1 database \cite{7358782}, the AIT database \cite{kauba2021towards} and the HDA database \cite{priesnitz2021mobile}. \rev{We consider three recognition workflows for the evaluation, one Commercial-Off-The-Shelf (COTS) system, one which is based on open-source algorithms and the NIST NBIS framework}. Our MCLFIQ model is benchmarked against the latest version of NFIQ 2 (NFIQ 2.2), a quality metric which is based on sharpness and was developed for contactless fingerprint images \rev{and the re-trained No-Reference Image Quality Assessment (NR-IQA) algorithm BRISQUE}. We report the predictive performance in terms of Error vs. Discard Characteristic (EDC) curves \cite{ISO-IEC-29794-1-QualityFramework-2023} for every combination of database, quality assessment algorithm and recognition workflow and analyze the EDC Partial Area Under Curve (EDC PAUC).
\end{itemize}

Our experiments show that a training of NFIQ 2 with a random forest classifier is possible and that synthetic data is a viable alternative to real databases. Our training results in the MCLFIQ model, which significantly outperforms all other methods in terms of predictive performance. Moreover, the MCLFIQ model shows a significantly improved robustness considering different databases and recognition workflows. Also, it is observable that sharpness is the most important quality measure for mobile contactless fingerprints. 
However, the amount of suitable contactless fingerprint databases is limited so that our method could only be tested on rather small databases. 
\rev{Based on our investigations, we suggest to further investigate on contactless fingerprint quality assessment and consider the proposed MCLFIQ method as a starting point  for a new standard algorithm.
For this reason, the MCLFIQ model is made publicly available so that interested researchers can download and test MCLFIQ on their own databases and benchmark it against NFIQ 2.2 or other methods. Furthermore, we will provide the pre-processing pipeline so that it can be used and refined for other databases.\footnote{The MCLFIQ model pre-processing pipeline will be made publicly available upon acceptance.}}

The rest of the paper is structured as follows: Section \ref{sec:related_work} discusses the related work. In Section \ref{sec:sample_quality} aspects of quality assessment for contactless fingerprint recognition are presented and the applicability of NFIQ 2 is evaluated. Section \ref{sec:proposed_system} introduces our proposed system. In Section \ref{sec:experimental_setip} the experimental setup is explained based on which the experimental results are summarized in Section \ref{sec:discussion}. Finally, Section \ref{sec:conclusion} concludes the paper. 

\section{Related work}
\label{sec:related_work}

Very few works investigate the sample quality of contactless fingerprint samples. Table \ref{tab:relatedwork} gives an overview of the most relevant works proposed so far. 

Parziale and Chen \cite{Parziale2009} proposed a coherence-based quality measurement. This approach measures the strength of the dominant direction in a local region. For this purpose, the authors applied a normalized coherence estimation on local gradients of the gray level intensity. Moreover, the covariance matrix of the gradient vectors was denoted, which represents the clarity of the ridge line structure. The algorithm is applied in different block sizes, whereas the individual results are averaged. The resulting "global quality index" is an open interval which is not normalized. On their test database, the authors tested the proposed quality algorithm and divided the histogram into three equal-sized groups. In experiments, it was shown that the accuracy on partitions assessed with high quality is better than on partitions assessed with lower quality. 

Labati \etal \cite{labati2010neural} implemented 45 different quality features which include, among others, the fingerprint Region Of Interest (ROI), various Fourier features and Gabor features. The authors studied different subsets of the implemented features and proposed the most promising feature vectors for their database. A feed-forward neural network and a k-Nearest-Neighbor (kNN) classifier were used to aggregate the individual feature vectors to a final quality score. In contrast to the aforementioned work, the authors achieved a closed interval quality scores. The authors used a rather constrained data set and compared their method to NFIQ 1.0 \cite{Tabassi-NFIQ1-NISTIR-7151-2004}. It was shown that their own approach performed significantly better than the NFIQ 1.0 algorithm. However, it remains unclear if these findings generalize.  

Li \etal \cite{li2013quality, li2013autocorrelation} introduced a quality assessment algorithm for finger images acquired with smartphones. The authors used different metrics in the spatial and frequency domain, which resulted in a feature vector. A Support Vector Machine (SVM) was trained to separate high-quality blocks from those with low quality. Like Parziale and Chen, the authors did not normalize the algorithm to achieve a closed interval of quality scores. They also grouped the considered database into three partitions according to quality scores. Again, the EER is lower on partitions of higher quality.

Liu \etal \cite{7566671} evaluated generic quality factors for different contactless modalities, including contactless fingerprints. They conclude that contrast, sharpness, luminance and artifacts like sensor noise or compression artifacts are the most important factors. This general assessment highly corresponds to the findings of the other authors. It can be seen that sharpness and contrast related features like Fourier transformations (\cf \cite{li2013quality, li2013autocorrelation, labati2010neural}), Gabor filters (\cf \cite{Parziale2009, labati2010neural}) or image entropy (\cf \cite{labati2010neural}) are most suitable. Also, using different block sizes appears to be promising for a robust assessment (\cf \cite{Parziale2009, labati2010neural}).

The capabilities of using the contact-based quality assessment algorithms NFIQ 1.0, NFIQ 2.0 and one which is included in the Veridium SDK were evaluated by Wild \etal \cite{wild2019comparative}. The authors used a self-acquired database to evaluate the algorithms. In their experiments, the authors filtered samples based on quality scores and report EERs on the left-over subsets. From their results, it is observable, that all algorithms showed the intended behavior of assigning low-quality values to samples which have a low biometric performance. The results also indicate that all algorithms might perform better on contact-based samples. 

A preliminary evaluation of NFIQ 2.0 on contactless fingerprint and contact-based fingerprint databases was conducted by Priesnitz \etal \cite{9211015}. The authors evaluated NFIQ 2.0 on publicly available data and reported its predictive power \rev{in terms of EDC curves}. The study indicates that NFIQ 2.0 is in general suitable for the assessment of contactless samples, but a proper pre-processing is crucial for a high predictive power. Howerver, the predictive performance of NFIQ 2.0 on contactless data is in general worse compared to the tested contact-based data. 
From that, we can conclude that NFIQ 2.0 in its current version is not suitable for contactless fingerprint quality assessment.

For contactless fingerprint recognition schemes, a major challenges are a narrow field of depth, which may cause a de-focused fingerprint region of interest, low-quality camera setups and a blur caused by a finger movement during the capturing attempt. Kauba \etal \cite{kauba2021towards} discussed a Canny filter-based quality assessment algorithm which analyses the sharpness.  Nevertheless, these methods have to be normalized in certain quality ranges and hardly generalize to new capturing schemes. 

From the literature review, we can observe several weaknesses in the state-of-the-art. Many proposals of new algorithms were only evaluated on one database, which is not publicly available. \rev{Also, from the evaluation methodology used in the previous works, no clear conclusion  regarding the predicitve performance of the suggested quality assessment method  can be extracted. Most works only report a higher biometric performance if a subset \eg above a certain quality threshold is considered. Futhermore, many of the proposed algorithms do not consider any fingerprint related features, instead they focus on sharpness or contrast measures. Therefore, it is assumed that sharp images with a high fidelity result in high quality scores, which does not correspond to utility because fingerprint character is neglected. 
Moreover, it is observed that NFIQ 2 shows many advantages over other proposed algorithms, \eg quality features which have proven beneficial for a robust quality assessment. Furthermore, the included random forest classifier can be re-trained for the special characterisics \eg caused by the capturing setup.  For this reason, NFIQ 2 is seen as the most promising framework for a proposal of a contactless fingerprint quality assessment.}


\section{Biometric Sample Quality for Contactless Fingerprint Images}
\label{sec:sample_quality}
This section introduces prerequisites and requirements for utility-driven quality assessment. Finally, the general suitability of features included in NFIQ 2 is evaluated. 


\subsection{Prerequisites for Quality Assessment}
\label{subsec:prereq}
State-of-the-art mobile contactless fingerprint recognition setups as proposed in \cite{kauba2021towards, priesnitz2021mobile, carney2017multi} typically implement an automatic workflow which captures four inner-hand fingers. The capturing subsystem has to ensure that the captured image fulfills certain prerequisites to be suitable for further processing. 

\begin{itemize}

    \item \textbf{Finger separation:} In a four finger capturing scenario, the finger ROIs have to be separated from each other. Here, the captured hand photo is separated into four individual finger images. This task can be done \eg by deep learning-based methods \cite{Priesnitz-Fingerprint-Segmentation-ICPR-2020} or an on-screen guidance \cite{carney2017multi}.
    
    \item \textbf{Size of the fingerprint ROI:} In mobile contactless capturing scenarios, the distance between the capturing device and the hand can be freely chosen. However, if the distance is too high, the resolution of the ROI is too low for an extraction of the ridge pattern. On-screen guidance and user feedback can effectively avoid this capturing failure. 

    \item \textbf{Brightness and contrast of the ROI:} Especially in capturing scenarios with unconstrained illumination, the capturing subsystem has to ensure that the captured ROI has a proper  brightness and contrast which allows an extraction of the ridge pattern. Here, an additional light source, \eg the flashlight of a smartphone, is considered to be beneficial. 

    \item \textbf{Sharpness:} Contactless capturing schemes are vulnerable to sharpness related issues caused by movement and lens properties. In capturing scenarios with low environmental light, the shutter speed is slow and therefore the captured image can contain motion blur. Furthermore, the aperture of the lens is very high, which leads to a very narrow depth of field. This can lead to de-focused images with no clear ridge pattern. 

    \item \rev{\textbf{Finger positioning:} In the unconstrained capturing setups the fingers can be presented to the camera at various angles. Here, yaw and pitch angles can be easily corrected, whereas rotations around the roll angle are a major challenge. A strong deviation from a central finger perspective leads to shifted minutiae position and hence degraded accuracy. Further, it introduces blurring, because of the narrow depth-of-field. }

\end{itemize}

\subsection{Applicability of NFIQ 2 for Contactless Fingerprints}
NFIQ 2 represents the de facto standard quality assessment algorithm for contact-based fingerprinting at a resolution of 500 ppi. 
\rev{
The random forest classifier in combination with hand-crafted quality components included in NFIQ 2 offers distinct advantages over deep-learning approaches, \eg CNN-based methods:
\begin{itemize}
    \item CNNs do not provide an easy way to give actionable feedback to the user. In contrast, with NFIQ 2, it is possible to make well-founded decisions based on a subset quality components. 
    \item CNNs tend to generalize poorly to new data. The goal of NFIQ 2 is to provide an algorithm that is suitable for all recognition workflows which it was designed for.
\end{itemize}
}

NFIQ 2 incorporates 74 unique hand-crafted quality components, which have a high predictive performance and comprehensively cover important aspects of fingerprint images, \eg minutiae count, contrast, sample clarity or size of the ROI. \rev{Furthermore, NFIQ 2 can provide actionable feedback based of individual features to the user \eg if a fingerprint sample blurred or lacks contrast.}

The NFIQ 2 feature vector 
also includes measures which are highly sensitive to the sharpness of a fingerprint sample. It uses a  linear regression function to determine a gray level threshold and classifies the pixels as ridge or valley. Afterward, the \textit{local clarity score} is computed as the block-wise clarity based on those ridges and valleys. The \textit{orientation certainty level} indicates whether a fingerprint sample contains a clear ridge-line structure or not. For this, the strength of the energy concentration along the ridge flow is analyzed. If a fingerprint sample is rather sharp, the orientation certainty level is higher, which subsequently indicates a higher utility. 
Also, the component \textit{ROI Orientation Map Coherence Sum} computes as features the coherence map of the orientation field estimation by analyzing oriented patterns as described in \cite{KASS1987362}. 
Many of the NFIQ 2-features are directly comparable to features which have been considered for contactless fingerprint quality assessments in previous works, \cf Section \ref{sec:related_work}. 
The entire list of features can be seen in Table \ref{tab:feature_importance_full}. 

Since many features included in NFIQ 2 accurately assess quality measures of contactless fingerprints, NFIQ 2 is in general applicable for contactless fingerprints. However, its predictive performance is degraded compared to its performance on contact-based fingerprints \cite{9211015}. This is because the random forest is not trained on contactless fingerprints. 

As discussed, contactless fingerprints present different challenges compared to contact-based ones. For this reason, the importance of each individual feature has to be adjusted to improve the predictive power. This can be achieved by re-training the random forest classifier of NFIQ 2.

\begin{figure}[!t]
	\centering
	\includegraphics[width=0.725\linewidth]{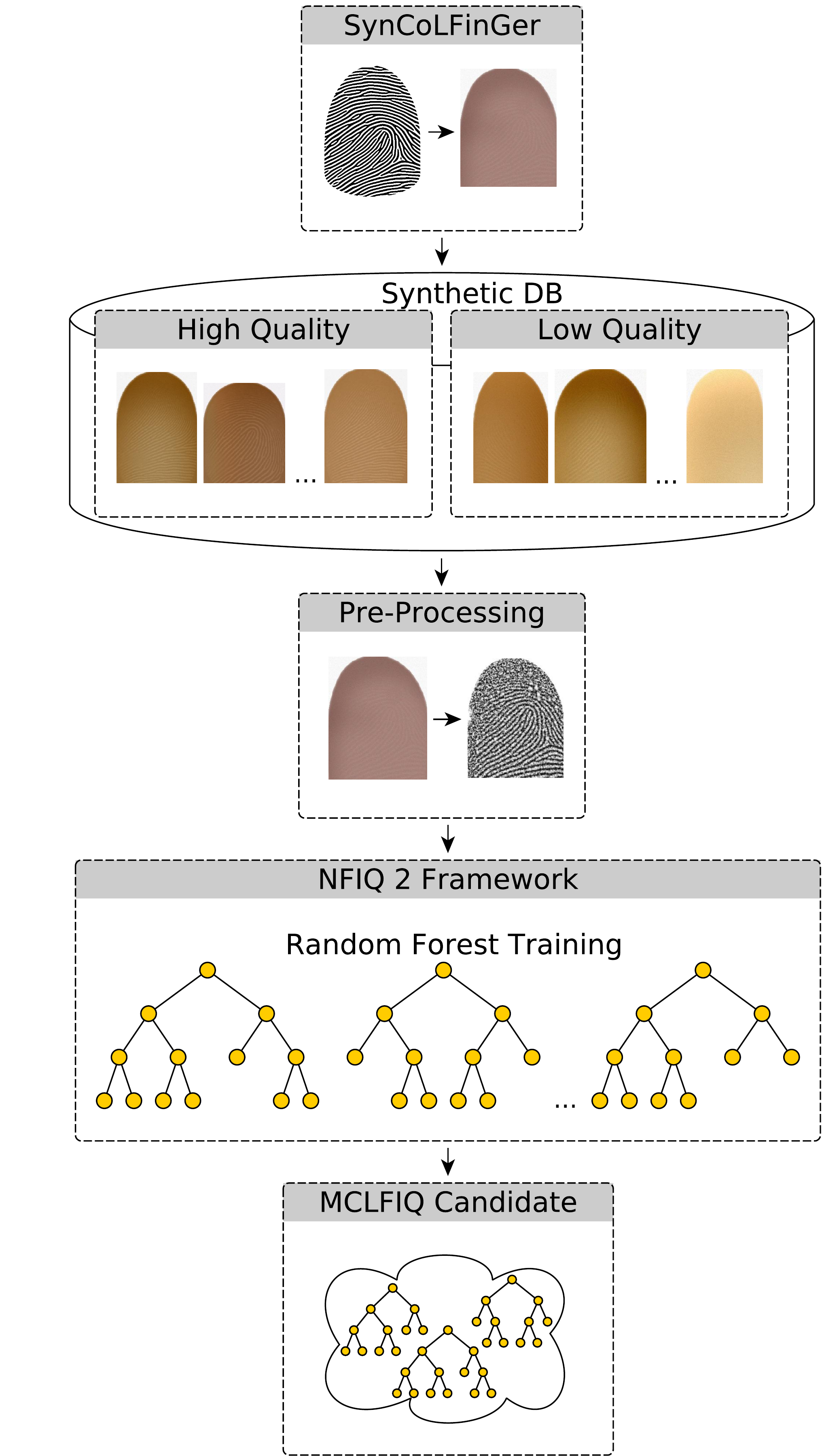}%
	\caption{Overview of the \review{MCLFIQ training  workflow using the NFIQ 2 framework} and important metrics for training and evaluation of the Random Forest (RF).}
	\label{fig:overview_training}
\end{figure}

For training NFIQ 2, annotated data is required. Since the training uses a binary classification algorithm, the training data has to be labelled with labels for \emph{high} and \emph{low} fingerprint quality.  Then, the training of the random forest consists of two steps: firstly, the feature extraction sub-system computes all features for the labeled training data and secondly, the random forest determines a configuration for every tree. 
From the final configuration, we can determine the importance of every feature, which indicates how much influence each single feature has on the final unified quality score. 

It should be noted, that \textit{NFIQ 2} refers to the general redesign as documented in NISTIR 8382 \cite{Tabassi-NFIQ21-NISTIR-8382-2021}, whereas \textit{NFIQ 2.2} refers to an individual release\footnote{https://github.com/usnistgov/NFIQ2}. NFIQ 2.2 was the latest version at the time the experiments have been conducted. 

Using NFIQ 2 as the basis for a contactless fingerprint quality assessment algorithm has several advantages.  We can use a set of well-engineered, tested and ISO/IEC 29794-4 compliant features which have been precisely calibrated on fingerprint samples \cite{ISO29794-4}. Thus, we are consistent with the vast majority of contactless fingerprint recognition schemes that use feature extraction algorithms from the contact-based domain.   



\section{The MCLFIQ Method}
\label{sec:proposed_system}

The NFIQ 2 framework is designed in a way that it is possible to adjust it to special characteristics of fingerprint images, \eg samples captured with different sensor types. Specifically, the random forest parameters are trained on data captured by the target capturing device type. Here, the included quality features and the range of quality scores remain the same, which makes different models highly comparable to each other. For this reason, the NFIQ 2 re-training framework is highly capable of proposing a new model for mobile contactless fingerprint images. 

\subsection{Sample Pre-processing}
\label{subsec:preprocessing_MCLFIQ}
Contactless finger images which fulfil the prerequisites discussed in \ref{subsec:prereq} are not automatically aligned to the requirements of quality assessment and recognition. For this reason, contactless fingerprint images need to be pre-processed in order to make them processable with established algorithms. A contactless fingerprint pre-processing pipeline is a set of algorithms which transfer the colored contactless finger image into a fingerprint sample. Many combinations of pre-processing algorithms show advantages and drawbacks on different databases. 
For this reason, we define framework requirements for a contactless fingerprint sample instead of specifying concrete algorithms:
\begin{itemize}
    \item \textbf{Rotation:} The fingerprint sample should be rotated to an upright position. 
    \item \textbf{Cropping:} The sample should only contain the fingerprint area.  Also, the fingerprint image should be cropped approximately at the first finger knuckle. 
    \item \textbf{Normalization:} The sample should be normalized to a ridge-line frequency of approximately \review{8 -- 12 pixels \cite{ISO29794-4}}.
    \item \textbf{Background separation:} The fingerprint sample should be precisely sepearated from the background so that the non-fingerprint area should appear in white. 
    \item \textbf{Gray scale conversion:} The fingerprint sample should contain only gray scale values. 
    \item \textbf{Emphasized ridge pattern:} The ridge pattern should be emphasized to align to a contact-based fingerprint. 
\end{itemize}

These requirements highly align with recommendations of many established tools from the contactless and contact-based domain. Also, many works propose pre-processing workflows which align to these requirements \cite{deb2018matching, wang2016preprocessing, hiew2007touch, sisodia2017conglomerate, grosz2021c2cl, jawade2021multi}. 
From the cited literature, it is also observable that many different capturing methods are used and that the pre-processing algorithms are optimized for the dedicated capturing setup.

\subsection{Training Process}
For the MCLFIQ training, we use the NFIQ 2 framework and benefit from its advantages of an established random forest framework and standardized quality features for fingerprint samples. Nevertheless, MCLFIQ is a unique model since we re-trained NFIQ 2 on an entirely different database and hence, changing its use case from contact-based to contactless fingerprints.  An overview of the conducted training steps can be seen in Figure \ref{fig:overview_training}.

The training process requires a data set of contactless fingerprint images which includes samples of high and low quality. To the author's knowledge, there are only a few contactless fingerprint databases publicly available, which are too small to train the NFIQ 2 random forest classifier. Also, the training and testing should be conducted on different databases, to give a fair indication of predictive performance. 

Moreover, labels which indicate whether a sample is of high or low quality are required. An algorithm for labeling biometric training data is proposed in \cite{ISO29794-4_2010}. However, this approach may be biased to the used comparison algorithm and may not be robust against miss-labeling, which negatively impacts the predictive performance of the proposed system. Alternatively, experts could label the fingerprints manually, which is time-consuming and requires an in-depth domain knowledge. For this reason, it is impractical to manually annotate large datasets.

Especially in the context of limited real data, using synthetic data for training the random forest classifier is an appropriate alternative \cite{Wood_2021_ICCV}. Another advantage is that  all available real-world databases remain available for testing purposes. We use the method proposed in \cite{PRIESNITZ2022127} for generating a mobile contactless fingerprint database for the training of MCLFIQ.
SynCoLFinGer is a synthetic contactless fingerprint generator which aims to generate samples captured by smartphones. SynCoLFinGer is based on a modelling approach which uses a SFinGe \cite{Cappelli2009} ridge pattern and applies various filters like deformations, distortions and noises to it which simulate a contactless capturing, subject characteristics such as skin color and environmental influences. For each filter, an intensity between 0 (low impact) and 100 (high impact) can be configured. Subsequently, the combination of all filter intensities defines the utility of the generated sample. For this reason, SynCoLFinGer can be precisely adapted to generate a well-suited training database of heterogeneous quality.

\begin{table*}[t]
\caption{Mobile contactless fingerprint databases considered for the evaluation. It should be noted, that the AIT database was captured in seven sessions under different environmental conditions.}
	\label{tab:dbs}
	\centering
    \scriptsize
		\begin{tabular}{cccccc}
			\hline
			\textbf{Database} & \textbf{Subset} & \textbf{Sensor} & \textbf{Capturing Setup} & \textbf{Instances} & \textbf{Samples} \\\hline
			\multirow{5}{*}{ISPFD} & NI (natural indoor) & Apple iPhone 5 & manual & 128 & 1,024\\
			& NO (natural outdoor) & Apple iPhone 5 & manual & 128 & 1,024 \\
			& WI (white indoor)& Apple iPhone 5 & manual & 128 & 1,024\\
			& W0 (white outdoor)& Apple iPhone 5 & manual & 128 & 1,024\\
            & LS (live scan) & Lumidigm Venus IP65 Shell & -- & 128 & 1,024\\ \hline
			\multirow{2}{*}{HDA} & const (constrained) & Google Pixel 4 & automatic & 28  & 448 \\
			& uncon (unconstrained) & Huawei P20 Pro & automatic  & 28 & 448\\
            & contact-based (cb) & CrossMatch Guardian 100 & -- & 29& 464 \\\hline
			\multirow{2}{*}{AIT} & mobile & Apple iPhone~11 & automatic & 14 & 1,568\\
            & contact-based (cb) & Green Bit DactyScan84c & -- & 10 & 1000 \\\hline
            \multirow{2}{*}{FVC2006} & DB2 & optical (BiometriKa) & -- & 150 & 1,800\\
             & DB3  & Thermal Sweep (Atmel) & -- & 150 & 1.800 \\\hline
		\end{tabular}	
\end{table*}



\section{Experimental Setup}
\label{sec:experimental_setip}
This section presents the experimental setup which is required to implement and evaluate MCLFIQ. First, the  training and evaluation databases are introduced. Further, algorithms for pre-processing and recognition are described. An overview of the considered evaluation workflow can be seen in Figure \ref{fig:overview_testing}.

\subsection{Training Database}
\label{subsec:training_database}
\begin{figure}[!t]
	\centering
    \begin{subfigure}[b]{0.225\textwidth}
    \includegraphics[width=0.4\linewidth]{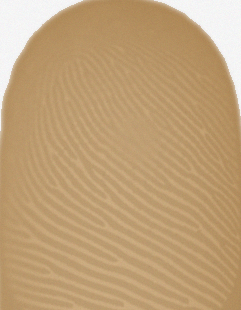}
    \hfil
    \includegraphics[width=0.4\linewidth]{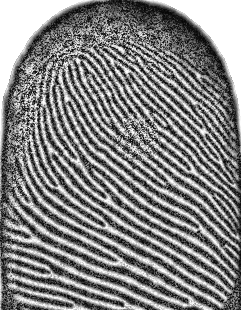}
    \caption{Quality preset: 87}
    \end{subfigure}
    \hfil
    \begin{subfigure}[b]{0.225\textwidth}
    \includegraphics[trim={0 1.5cm 0 0},clip,width=0.4\linewidth]{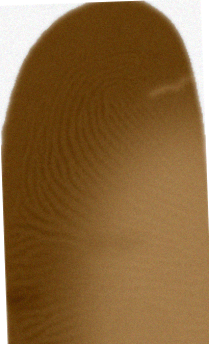}
    \hfil
    \includegraphics[trim={0 1.5cm 0 0},clip,width=0.4\linewidth]{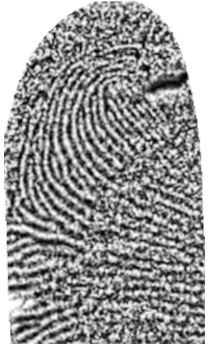}
    \caption{Quality preset: 18}
    \end{subfigure}
	\caption{Example images of the training database generated by SynCoLFinGer with corresponding pre-processed images and quality parameters: (a) high-quality preset and low (b) low-quality preset.}
	\label{fig:example_SynCoLFinGer}
\end{figure}
\review{

Due to the lack of publicly available databases, the contactless fingerprint training database considered in the work is generated synthetically by the SynCoLFinGer method \cite{PRIESNITZ2022127}. 
SynCoLFinGer includes the generation of contactless finger images of distinct quality. For this, the generation can be configured with a simple configuration parameter between 0 (low quality) and 100 (high quality). From this configuration parameter, the parameters for subject characteristics and environmental influences are derived. It should be noted that this configuration parameter is not correlated to NFIQ 2 scores. 

From a set of ridge-patterns, we generate two subsets: one of high quality and one of low quality. For this, the configuration parameters for the low-quality subset are set to a range between 0 and 33 whereas the high-quality subset is generated with a preset of values between 66 and 100. Figure \ref{fig:example_SynCoLFinGer} presents sample images of high and low quality, which are included in the training database. As can be seen from the picture, high-quality samples contain \eg less rotation distortions and a clearer ridge-line pattern, whereas a low-quality samples are distorted more and show more noise and dirt artifacts. It should be noted that due to random variables which are incorporated in SynCoLFinGer, the training database also contains samples of moderate quality. 
This method automatically generates the ground truth by assigning high and low quality labels to the samples. 
}

\subsection{Evaluation Databases}

\begin{figure*}[!t]
	\centering
    \begin{subfigure}[b]{0.2\textwidth}
    \centering
    \includegraphics[width=0.99\linewidth]{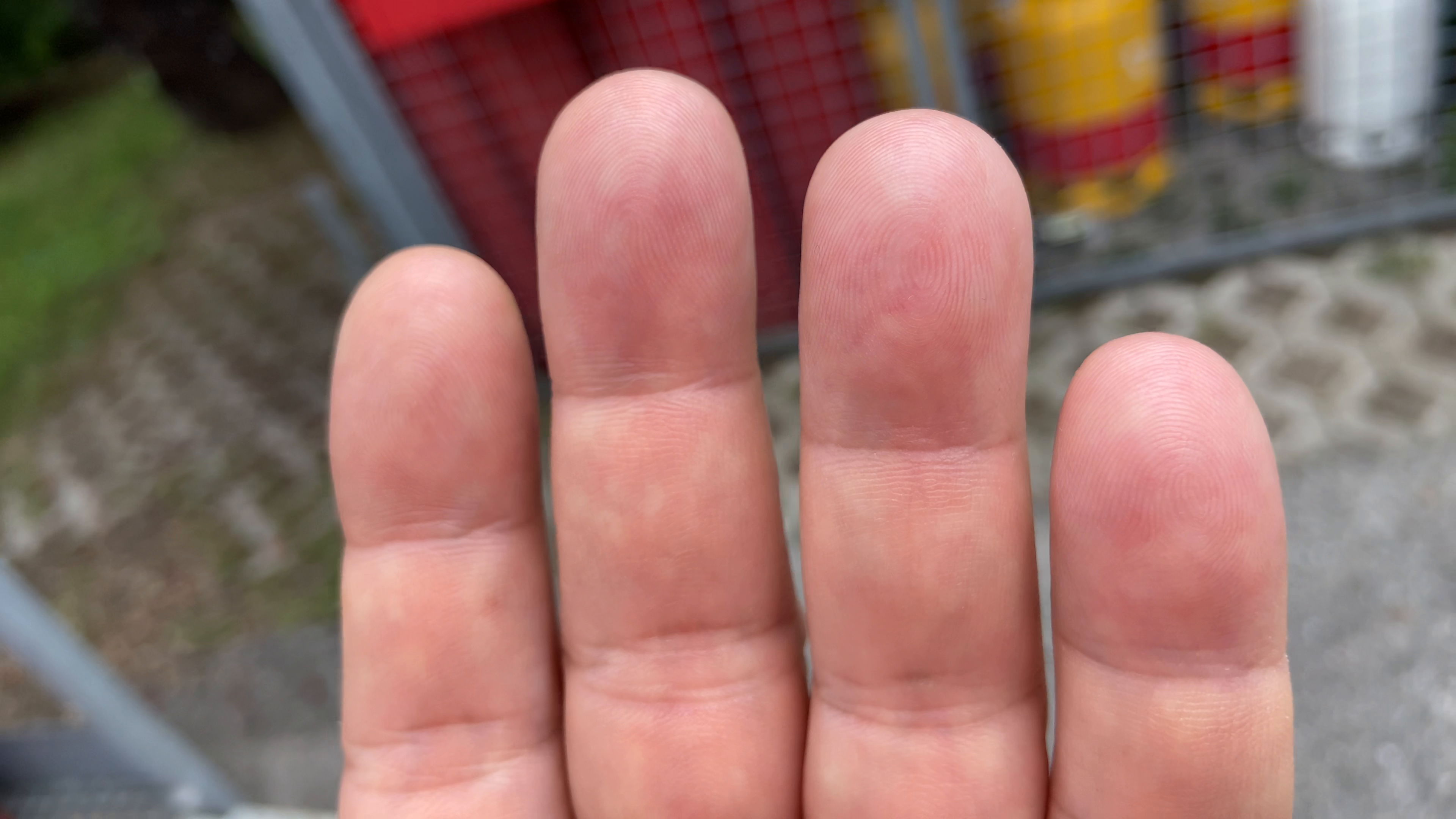}
    \caption{}
    \end{subfigure}\hfil
    \begin{subfigure}[b]{0.4\textwidth}
    \includegraphics[width=0.17\linewidth]{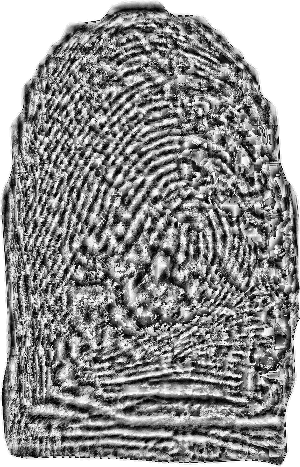}\hfil
	\includegraphics[width=0.17\linewidth]{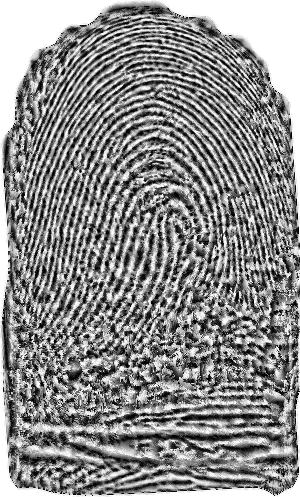}\hfil
	\includegraphics[width=0.17\linewidth]{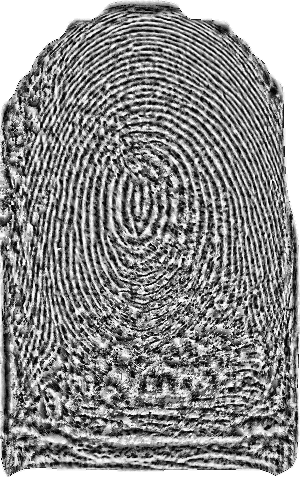}\hfil
	\includegraphics[width=0.17\linewidth]{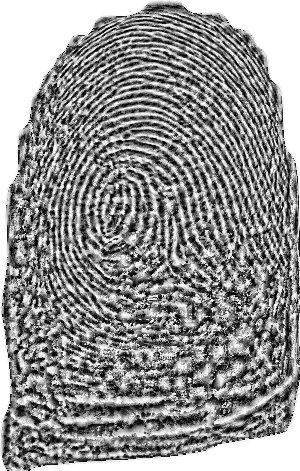}
    \caption{}
    \end{subfigure}
    \hfil
    \begin{subfigure}[b]{0.2\textwidth}
    \centering
	\includegraphics[width=0.375\linewidth]{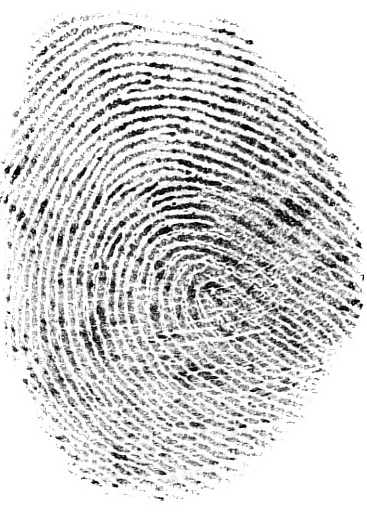}
     \caption{}
    \end{subfigure}\hfil
 	\caption{Example images of the AIT mobile database captured in scenario 6 (lattice): (a) before pre-processing, (b) after pre-processing, (c) contact-based.}
    \label{fig:example_AIT}
    \end{figure*}

\begin{figure*}[!t]
	\centering
     \begin{subfigure}[b]{0.2\textwidth}
     \centering
	\includegraphics[width=0.99\linewidth]{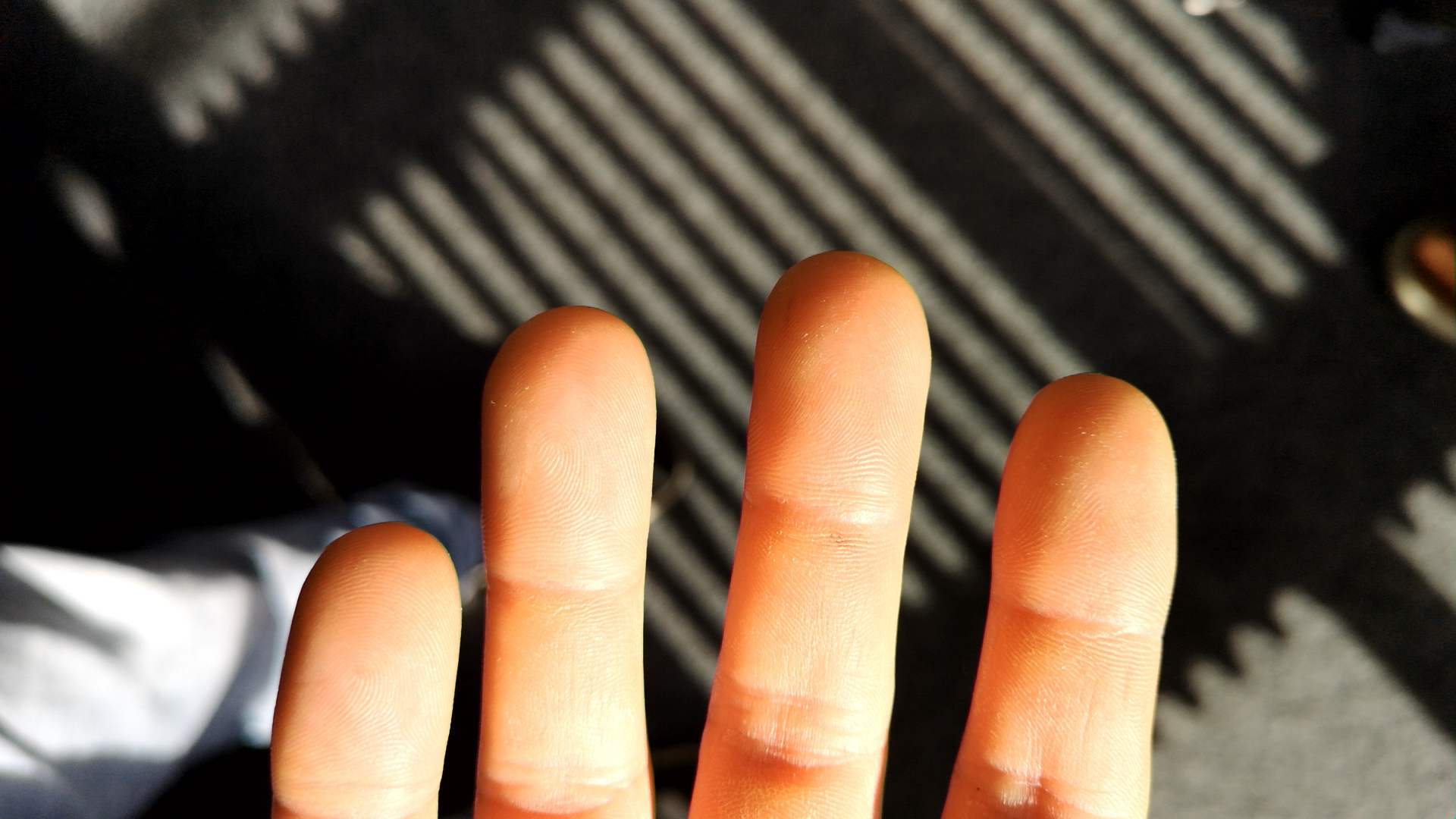}
     \caption{}
    \end{subfigure}
    \hfil
    \begin{subfigure}[b]{0.4\textwidth}
    \centering
	\includegraphics[width=0.17\linewidth]{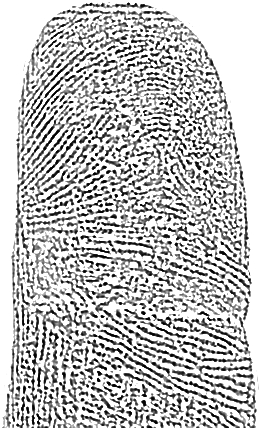}\hfil
	\includegraphics[width=0.17\linewidth]{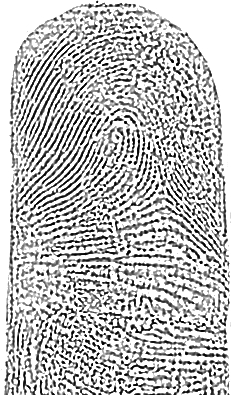}\hfil
	\includegraphics[width=0.17\linewidth]{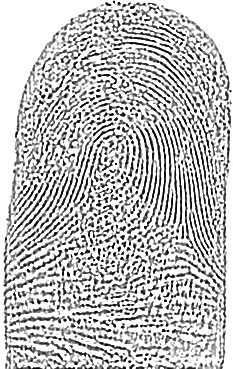}\hfil
	\includegraphics[width=0.17\linewidth]{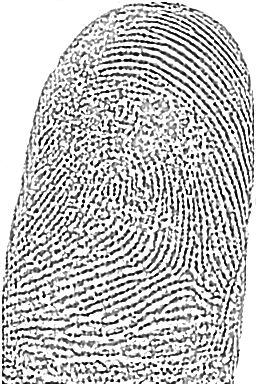}
    \caption{}
    \end{subfigure}
    \hfil
    \begin{subfigure}[b]{0.2\textwidth}
    \centering
	\includegraphics[width=0.375\linewidth]{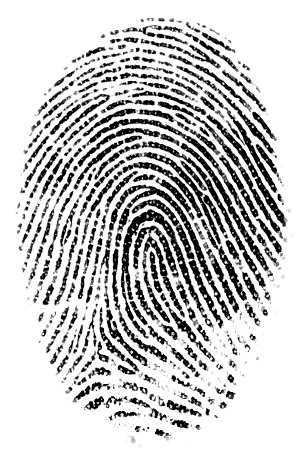}
     \caption{}
    \end{subfigure}
	\caption{Example images of the HDA database taken from the unconstrained capturing scenario: (a) before pre-processing, (b) after pre-processing, (c) contact-based.}
	\label{fig:example_HDA}
\end{figure*}
\begin{figure}[!t]
	\centering
    \begin{subfigure}[b]{0.2\textwidth}
	\includegraphics[width=0.85\linewidth]{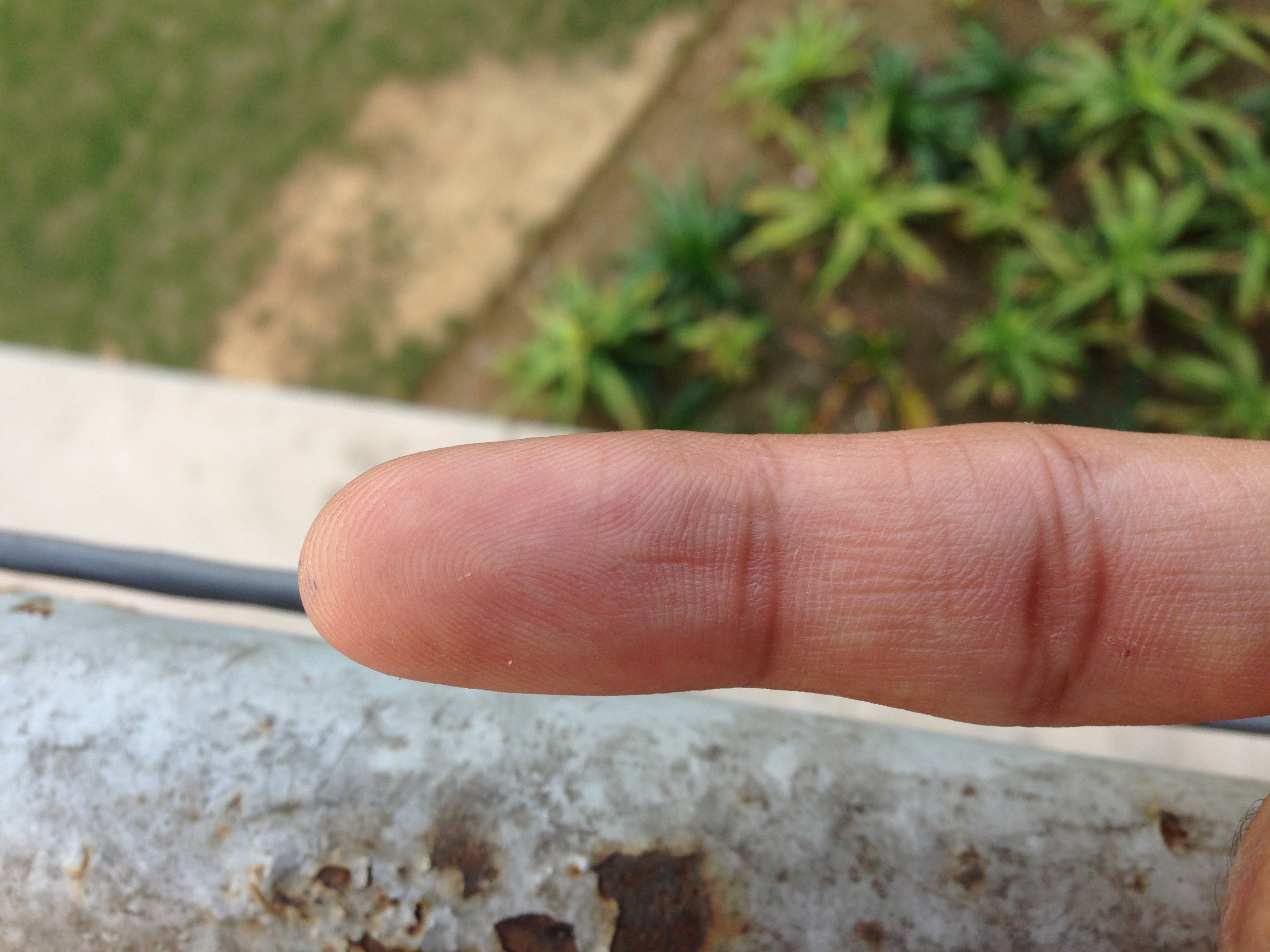}
    \caption{}
    \end{subfigure}
    \hfil
    \begin{subfigure}[b]{0.1\textwidth}
	\includegraphics[width=0.75\linewidth]{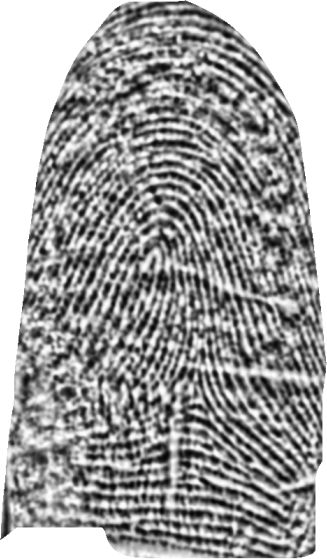}
    \caption{}
    \end{subfigure}
    \hfil
    \begin{subfigure}[b]{0.1\textwidth}
	\includegraphics[width=0.9\linewidth]{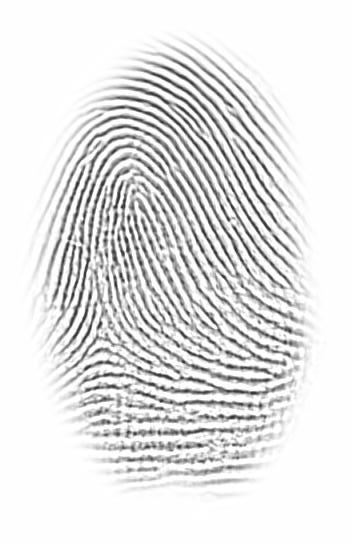}
    \caption{}
    \end{subfigure}
	\caption{Example images of the ISPFD database taken from the natural outdoor sub-database: (a) before pre-processing, (b) after pre-processing, (c) contact-based.}
	\label{fig:example_iiitd}
\end{figure}

For our experiments, we employ three different contactless fingerprint evaluation databases. All databases are captured using smartphones in a mobile scenario. Used databases and their properties are listed in Table~\ref{tab:dbs} and briefly summarized as follows:

	\textit{AIT Database}~\cite{kauba2021towards}:
    The dataset consists of 14 subjects 
    whose four inner hand fingers 
    were recorded in 7 different scenarios. The scenarios consist of two office-like environments, four open-air scenarios and one cellar scenario to simulate nighttime recording. The acquisition was carried out using an iPhone~11, which recorded videos.    In total, 196~videos were recorded.
    Each video with a duration between 10 and 15 seconds was split in two parts of equal duration. For each video part, the fingerprint with the highest sharpness within the first five seconds was selected and extracted. As a result, the dataset is composed of 1,568 contactless fingerprint samples in total. \rev{The database also contains a subset of contact-based samples.} Example images of the database are presented in Figure \ref{fig:example_AIT}. Further details about the dataset can be found in~\cite{kauba2021towards}. 

	\begin{figure}[!t]
	\centering
    \begin{subfigure}[b]{0.22\textwidth}
    \centering
	\includegraphics[width=0.45\linewidth]{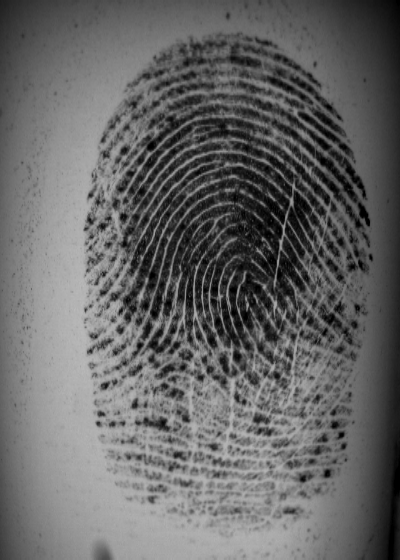}
    \caption{}
    \end{subfigure}
    \hfil
    \begin{subfigure}[b]{0.22\textwidth}
    \centering
	\includegraphics[width=0.525\linewidth]{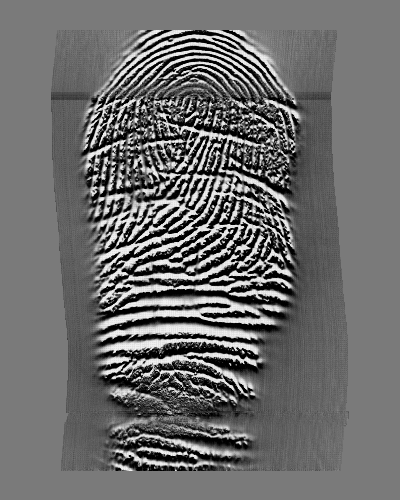}
    \caption{}
    \end{subfigure}
	\caption{\rev{Examples of the FVC2006 database: (a) DB2, (b) DB3.}}
	\label{fig:example_fvc}
\end{figure}

	\textit{HDA Database} \cite{priesnitz2021mobile}: The HDA database consists of contactless samples captured in two different setups. A box-setup simulates a constrained dark environment, whereas a tripod-setup simulates an unconstrained capturing scenario.
	For the capturing, we used two different smartphones: the Google Pixel 4 (constrained scenario) and the Huawei P20 Pro (unconstrained scenario). An application automatically captured the four inner-hand fingers and processed them to fingerprint samples. \rev{During the database acquisition, also contact-based samples were captured.} Example images of the database are presented in Figure \ref{fig:example_HDA}.
    
	\textit{ISPFD} \cite{7358782}: the IIITD SmartPhone Fingerphoto Database v1 consists of contactless fingerprint images acquired using an Apple iPhone 5 smartphone. It includes finger images captured in indoor and outdoor scenarios with natural and white background. Figure~\ref{fig:example_iiitd} depicts example images of the ISPFD database. 

 	\rev{\textit{FVC2006} \cite{CAPPELLI20077}: the database of the fourth international Fingerprint Verification Competition (FVC), contains four disjoint fingerprint subsets. The first three subsets are each collected with a different contact-based sensor while the fourth database is synthetically generated. We only use subsets DB2 and DB3 as the others are not considered useful for our experiments. Example images of the FVC06 database are depicted in Figure~\ref{fig:example_fvc}.}

 \rev{It should be noted, that all considered contactless databases fulfill the prerequisites discussed in Section \ref{subsec:prereq}.
However, they are rather small compared to real-world application scenarios and do not represent  a typical population. }

\begin{table}[!t]
    \caption{Overview on the evaluation process including relevant metrics.}
    \centering
    \scriptsize
    \begin{tabular}{ccc}
    	\hline
    	\textbf{Quality Assessment} & \textbf{\makecell{Feature Extraction \\ and Comparison}} & \textbf{Evaluation Metric} \\
    	\hline
    	MCLFIQ &     \multirow{1.5}{*}{FingerNet and SourceAFIS} & \multirow{2}{*}{EDC} \\
    	NFIQ 2.2 & \multirow{2.25}{*}{IDKit SDK (Innovatrics)} &  \\
    	AIT Sharpness & \multirow{3}{*}{NBIS} & \multirow{2}{*}{DET} \\
        BRISQUE &&\\
    	\hline
    \end{tabular}
	\label{fig:overview_testing}
\end{table}

\subsection{Database Pre-processing}
\label{subsec:database_preprocessing}
To extract features from contactless fingerprints with tools designed for the contact-based domain, a pre-processing has to be applied which transfers a contactless finger image to a contact-based equivalent sample.
According to our suggestion in Section~\ref{subsec:preprocessing_MCLFIQ}, we use the same pre-processing pipeline for all databases to achieve a consistent impression on all samples. 

\rev{Since the ISPFD database contains unsegmented and un-rotated finger photos, the fingerprint region of interest is segmented by a deep-learning-based semantic method \cite{Priesnitz-Fingerprint-Segmentation-ICPR-2020}. The method uses a DeepLabv3+ model which was fine-tuned for the segmentation of fingertips.} The segmented finger image is then rotated to an upright position. All other databases provide already segmented and rotated fingerprint images, so this step is omitted.

\rev{The segmented data is then converted to gray scale and a Contrast Limited Adaptive Histogram Equalization (CLAHE) is applied to emphasize the ridge-line characteristics. The CLAHE algorithm is iteratively applied with decreasing size of tile grids. This process first equalizes the brightnes throughout the fingerprint region and second emphasizes the ridge pattern.}  Next, the fingerprint samples are normalized to a fixed ridge-line frequency of approximately 9\,pixels, which aligns to approximately 500\,ppi live-scanned fingerprints and is favored by NFIQ 2 and the recognition algorithms. \rev{Here, the ridge-to-ridge distance is measured and the fingerprint image is re-sized accordingly.} 
All images are converted into a uniform file format to fulfil the requirements of NFIQ 2 and the recognition workflows.

\subsection{Training Process}
The NFIQ 2 training framework is maintained by the International Organization for Standardization (ISO). The framework provides a process, which consists of steps for data labelling, training the random forest parameters and an evaluation of the random forest.

The labeled training database of our final training attempt consists of 40,000 synthetic samples, 30,000 for training and 10,000 for evaluation. The database is generated to consist of 50\% high-quality samples and 50\% low-quality samples (\cf Section~\ref{subsec:training_database}). 

During the training process, a new random forest is built based on the labeled training data. The training parameters (100 trees in the random forest, maximum depth of each tree of 25, 10 randomly sampled variables as split candidates, minimum sample count per leave of  2 and tree pruning) are the same as in NFIQ 2. During the validation process, the automatic assignment of quality labels has proven to work accurately, since only six samples have been miss-classified. These samples were generated with the high-quality preset, but validated by the NFIQ 2 framework as low quality. We manually re-labeled them and thus got a final training database of 19,994 high-quality samples and 20,006 low-quality samples.

The trained random forest outputs a class membership along with its probability. The final NFIQ 2 score is the probability that a given image belongs to class 1 multiplied by 100 and rounded to its closest integer.

\subsection{Considered Baseline Algorithms} 
To evaluate the predictive performance of MCLFIQ in comparison to established quality assessment algorithms we select NFIQ 2.2, a sharpness-based quality estimation algorithm introduced by the AIT \cite{kauba2021towards} and BRISQUE \cite{mittal2012brisque}.  

As discussed in the Section \ref{sec:related_work}, it is also possible to assess the quality of contactless fingerprints using NFIQ 2. Even though it is not designed for this use case, it includes many quality features which are also of high relevance for contactless samples. Additionally, the practical applicability on contactless fingerprints has been shown in \cite{9211015}.

\begin{figure}[!t]
	\centering
    \begin{subfigure}[b]{0.115\textwidth}
    \includegraphics[width=0.9\linewidth]{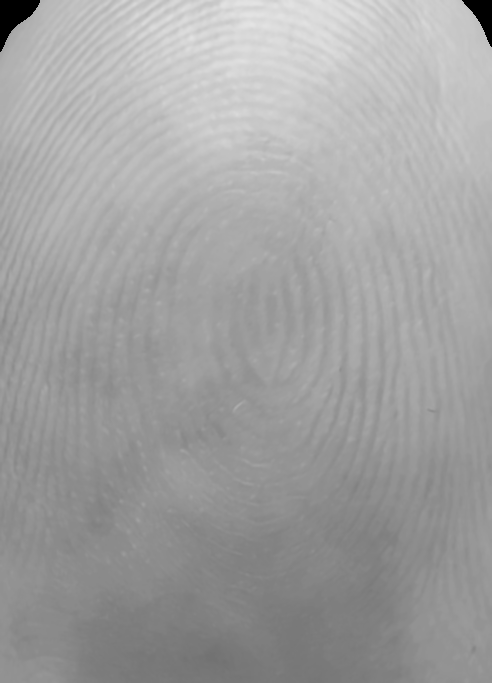}%
    \caption{}
    \label{fig_roi}
    \end{subfigure}
	\hfil
	\begin{subfigure}[b]{0.115\textwidth}
    \includegraphics[width=0.9\linewidth]{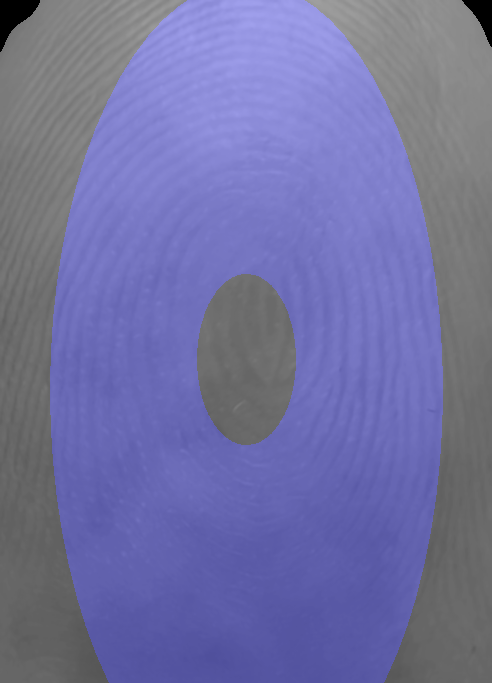}%
    \caption{}
	\label{fig_roi_mask}
    \end{subfigure}
	\hfil
	\begin{subfigure}[b]{0.115\textwidth}
    \includegraphics[width=0.9\linewidth]{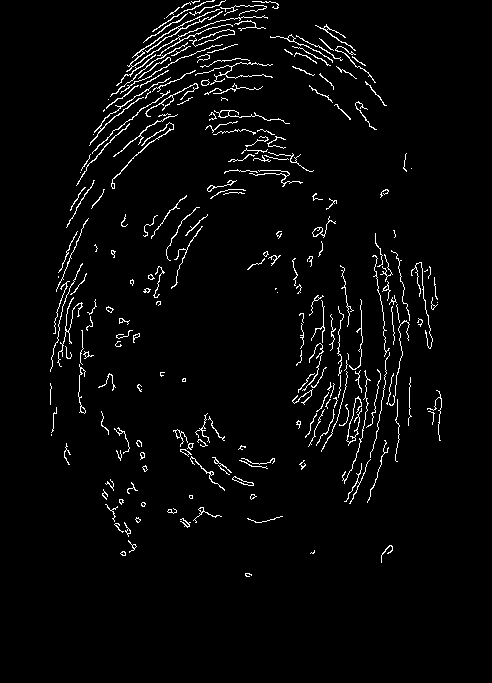}%
    \caption{}
 	\label{fig_third_case}

 \end{subfigure}
	\caption{Visualization of the sharpness-based quality estimation. (a) original fingerprint image (converted to gray scale), (b) superimposed elliptical mask, (c) computed Canny edges.}
	\label{fig_sim}
\end{figure}

As a second algorithm, we adapted a sharpness-based quality assessment algorithm introduced by Kauba \etal~\cite{kauba2021towards}. It works as follows: firstly, all fingerprints are scaled to the same image width in order to reduce the effect of the distance between fingertip and camera sensor on the sharpness calculation (Figure~\ref{fig_roi}). Secondly, an elliptical mask overlays the fingerprint image sample. The mask consists of two nested ellipses, and only the area between both ellipses is considered for calculating the sharpness (Figure~\ref{fig_roi_mask}). Thirdly, the Canny edge detector~\cite{canny1986computational} is applied for edge detection (Figure~\ref{fig_third_case}). Finally, the sharpness value is the ratio of the number of summed edges and the size of valid pixels as defined by the mask. We normalize the resulting floating-point sharpness value to an integer  between 0 and 100 in order to integrate it into our workflow. 

\rev{As a third quality assessment algorithm, we employ the blind/referenceless image spatial quality
evaluator (BRISQUE) introduced by Mittal \etal~\cite{mittal2012brisque}. BRISQUE is a no-reference image quality assessment algorithm designed to evaluate the naturalness and quality of images. 
The classification task is done by a Support Vector Machine (SVM) Regressor (SVR). We re-trained BRISQUE using the same SynCoLFinGer database, like for MCLFIQ. Like for MCLFIQ, the quality annotations in a range of [0, 100] of the training data are directly generated by SynCoLFinGer. It should be noted that no pre-porcessing, \ie no gray-scale conversion was conducted for the training data. Accordingly, for testing the samples were only segmented, cropped and rotated.}

\subsection{Recognition Algorithms}
For our experiments, we use three recognition algorithms, a Commercial Off-The-Shelf (COTS) system and two open-source fingerprint recognition systems. 
\paragraph{Commercial Off-The-Shelf (COTS) System}
\label{sec:COTS}
The fingerprint recognition system IDKit SDK from Innovatrics\footnote{IDKit SDK version 8.0.1.50, see https://www.innovatrics.com} is used as COTS software. The system is originally designed for contact-based fingerprint samples, but has also proven to work robustly with pre-processed contactless samples \cite{kauba2021towards}.

\paragraph{Open-Source Fingerprint Recognition System}
\label{sec:open_source}
The first considered open-source fingerprint recognition system is based on the FingerNet feature extractor of Tang \etal \cite{tang2017fingernet} and a minutiae pairing and scoring algorithm of the SourceAFIS system of Va\v{z}an \cite{Vazan-SourceAFIS-2019}. The original algorithm uses minutiae quadruplets, \ie additionally considers the minutiae type (ridge ending or bifurcation). Since minutiae triplets are extracted by the used minutiae extractors, the algorithm has been modified to ignore the type information since the SourceAFIS system does not support this information.

\paragraph{NIST NBIS Framework}
\rev{The second considered open-source fingerprint recognition system is the NBIS framework\footnote{NBIS version 5.0.0, see https://www.nist.gov/services-resources/software/nist-biometric-image-software-nbis} developed by NIST. We used the MINDTCT tool to extract minutiae information and BOZORTH3 to compute the template comparison scores. It should be noted that all experiments were conducted using the default parameters without any optimizations like minutiae quality threshold adaptions. }

\subsection{Measuring Biometric Sample Quality}
\label{subsec:measuring_biometric_performance} 
The aspects (\cf Section \ref{sec:intor}) of biometric quality have to be expressed in an objective manner to ensure that performance can be measured and compared between different systems.
Tabassi \etal \cite{Tabassi-NFIQ1-NISTIR-7151-2004} proposed an approach for objective performance assessment based on a measure of the distance between the mated and non-mated comparison score distributions for a given sample. Well separated distributions imply that the likelihood of false accept or false reject is low, and that it increases with greater overlap between the distributions. This approach is generalized in ISO/IEC 29794-1:2009 \cite{ISO-IEC-29794-1-QualityFramework-090205} which requires that the quality score output of a biometric quality assessment algorithm conveys the predicted utility of the biometric sample.

For evaluating the predictive power of a quality assessment algorithm of a biometric recognition system, Grother and Tabassi \cite{Grother-SampleQualityMetricERC-PAMI-2007} introduced the Error vs. Reject Curve (ERC). This method evaluates whether a rejection of low quality samples results in a reduced False-Non-Match error Rate (FNMR). 
 Each mated comparison is associated with a similarity score $s_{i}$ and two quality scores $q _i ^{(1)}$ and $q _i ^{(2)}$. In order to aggregate the pair of quality scores from a pair of samples to be compared, the $\min$ function is chosen as combination function:

\begin{equation}
\label{fmin}
q_i = \min\Big(q _i ^{(1)}, q _i ^{(2)}\Big)
\end{equation}

Then a set $R(u)$ is formed containing the pairwise minima which are less than a fixed threshold of acceptable quality $u$:

\begin{equation}
R(u) = \Big\{i:min\Big(q _i ^{(1)}, q _i ^{(2)}\Big) < u \Big\}
\end{equation}

Subsequently, $R(u)$ is used to exclude comparison scores and to compute the FNMR on the rest. Starting with the lowest of the pairwise minima, comparisons are excluded up to a threshold $t$ which is obtained by using the empirical cumulative distribution function of the comparison scores, which corresponds to a FNMR of interest denoted by $f$: 

\begin{equation}
\label{fexcl}
t = M^{-1}(1-f)
\end{equation}
The ERC is then computed by iteratively excluding a portion of samples and recomputing the FNMR on the remaining comparison scores which are below the threshold:

\begin{equation}
\mbox{FNMR}(t,u) = \frac{\big|\{s_{ii}:s_{ii} \le t,i \notin R(u)\}\big|}{\big|\{s_{ii}:s_{ii} \le \infty\}\big|}
\end{equation}

Due to the effect that a fraction of low-quality samples are excluded in every iteration step, the FNMR should decrease constantly if the quality measure is a good predictor for the biometric performance. 
This method is widely adopted by the research community and is also known as Error vs. Discard Characteristic (EDC) curve \cite{ISO-IEC-29794-1-QualityFramework-2023}.

In order to compare different EDCs, the area under each curve is computed  up to a pre-defined discard rate and denoted as Error vs. Discard Characteristic Partial Area Under Curve (EDC PAUC). Here, the threshold is set to $x = 0.2$ to only consider the most relevant part of the curve.



\section{Results}
\label{sec:discussion}

\begin{table*}
\caption{Overview on the ten most important features of MCLFIQ and \rev{the original} NFIQ 2.2 incl.  their relative importance.}
	\centering
 \scriptsize

	\begin{tabular}{L{0.275\linewidth}L{0.15\linewidth}|L{0.275\linewidth}L{0.15\linewidth}}
		\toprule
        \multicolumn{2}{c|}{\textbf{MCLFIQ}}
        &\multicolumn{2}{c}{\textbf{NFIQ 2.2}}
        \\
        \textbf{Feature name}
		& \textbf{Feature Importance (\%)}
		& \textbf{Feature name}
		& \textbf{Feature Importance (\%)}
		\\
		\midrule
		ROI Relative Orientation Map Coherence Sum 
		& 38.56
		& Frequency Domain Analysis Standard Deviation
		& 6.72
		\\
		Orientation Certainty Level Mean 
		& 21.29
		& FingerJet FX OSE COM Minutiae Count 
		& 4.40
		\\
		Orientation Certainty Level Bin 0 
		& 7.6
		& FingerJet FX OSE OCL Minutiae Quality 
		& 3.96
		\\
		ROI Orientation Map Coherence Sum 
		& 7.45
		& Ridge Valley Uniformity\_Mean 
		& 3.32
		\\
		FingerJet FX OSE OCL Minutiae Quality 
		& 5.46
		& Frequency Domain Analysis Mean 
		& 2.97
		\\
		Orientation Certainty Level Bin 7 
		& 5.27
		& FingerJet FX OSE Total Minutiae Count 
		& 2.75
		\\
		Frequency Domain Analysis Bin 0 
		& 2.13
		& Ridge Valley Uniformity Standard Deviation 
		& 2.43
		\\
		Orientation Certainty Level Bin 8 
		& 1.56
		& Local Clarity Score Bin 7 
		& 2.42
		\\
		Orientation Certainty Level Bin 6 
		& 1.29
		& Local Clarity Score Bin 8 
		& 2.39
		\\
		Frequency Domain Analysis Mean 
		& 1.07
		& Frequency Domain Analysis Bin 9 
		& 2.28
		\\
		\midrule
		\textbf{Sum} & \textbf{91.68}
		& \textbf{Sum} & \textbf{33.65}
		\\
		\bottomrule
	\end{tabular}
	
    \label{tab:feature_importance}
\end{table*}

In this section, we present the results of the MCLFIQ  training and validation.  Furthermore, we discuss the evaluation in terms of biometric performance and predictive power. 

\subsection{Training and Validation Results}
Two important identifiers for the  training accuracy are the training error rate and the out-of-bag error. The training error shows the number of samples which cannot be predicted correctly according to their ground truth labels. 
The out-of-bag error defines the mean prediction error averaged over each training sample, using only the trees that did not have the sample in their bootstrap. 

We aim to align our training and validation results to the results of the original NFIQ 2.2 training. For this reason, we designed our final training database to have zero training error and a low out-of-bag error of 0.0009. In comparison, NFIQ 2.2 has also a training error of zero and an out-of-bag error of 0.24 \cite{Tabassi-NFIQ21-NISTIR-8382-2021}. 

The validation determines how the random forest generalizes to unseen data. Here, the validation error rate shows how many samples are miss-predicted according to their labels. The validation error rate is 0 for both models. As discussed, we optimized the training database to achieve the same results as in the original NFIQ 2.2 model.

\rev{During the random forest training process the importance, of every individual feature is adjusted. This means that during the training process, it is evaluated which feature has a high share of a correct attribution of a quality score and which does not. 
The overview of the relative importance over all feature included in the method indicates which features have highest relevance for the quality assessment. }
\begin{figure*}[!t]
	\centering
	\subfloat[Contactless]{\includegraphics[width=0.375\linewidth]{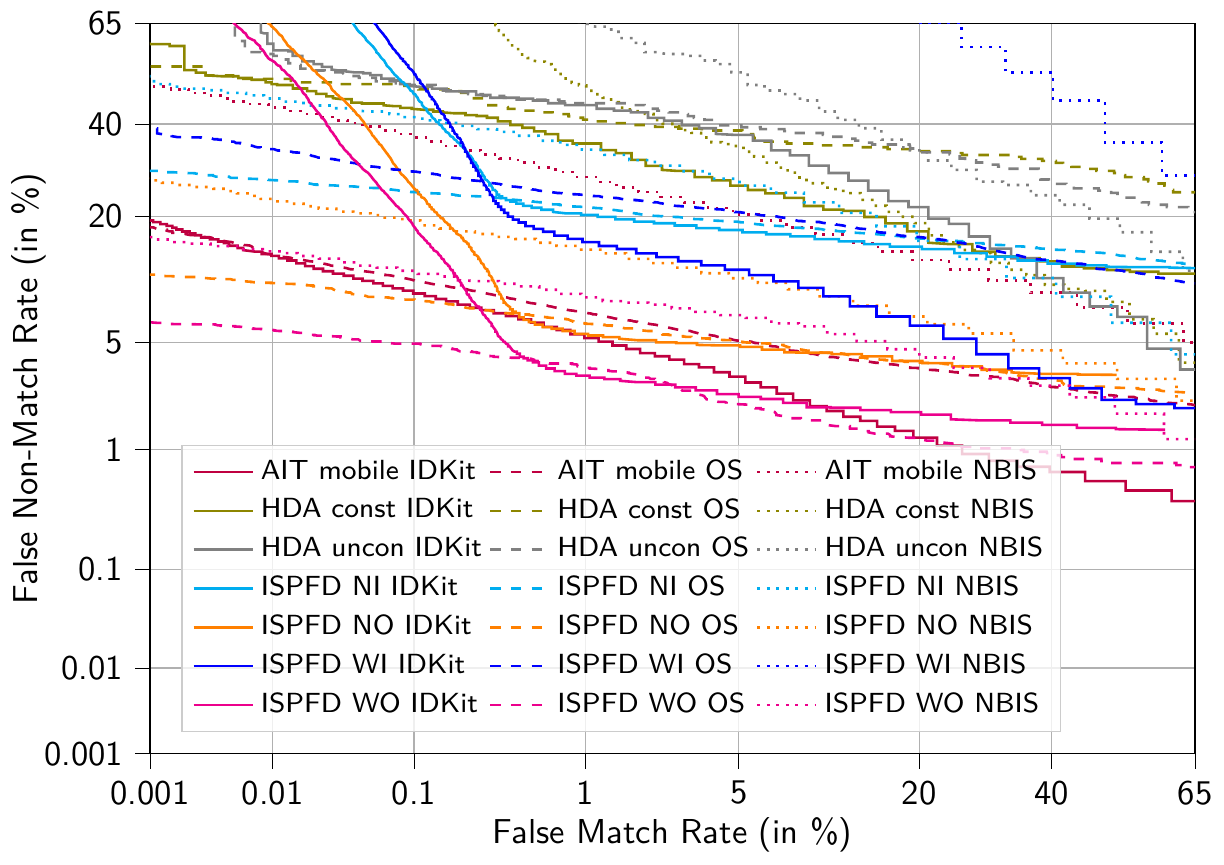}
	\label{fig:DET_AIT}}
	\hfil
	\subfloat[Contact-based]{\includegraphics[width=0.375\linewidth]{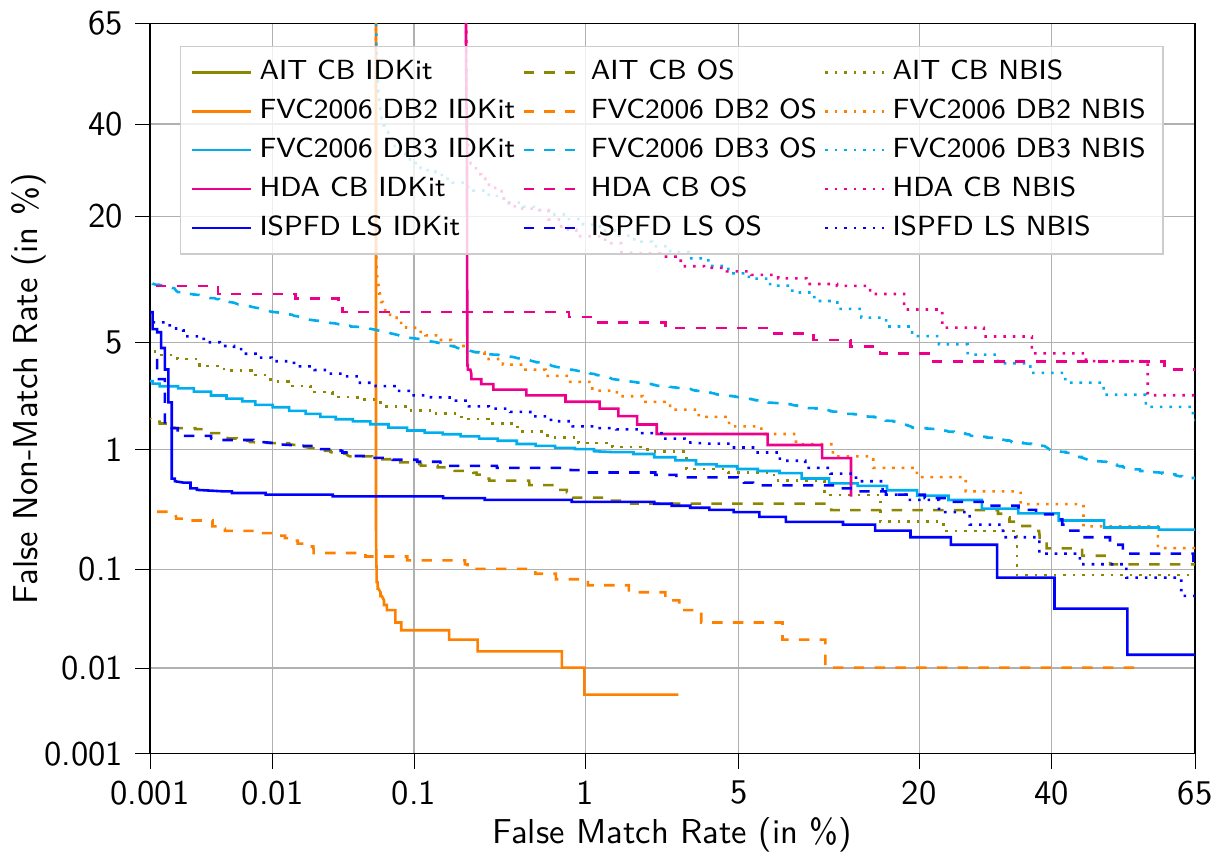}
	\label{fig:DET_HDA}}
	\caption{DET curves obtained on the considered databases using all recognition workflows. 
 }
	\label{fig:det}
\end{figure*}
Table \ref{tab:feature_importance} presents the 10 most important features of NFIQ 2.2 and MCLFIQ. We can observe that NFIQ 2.2 has a more uniformly distributed feature importance, whereas the MCLFIQ model relies mainly on a few features which have a high importance. In particular, the two features \textit{ROI Relative Orientation Map Coherence Sum} and \textit{Orientation Certainty Level Mean} combined share over 50\% of the whole feature importance. Both features are mainly based on sharpness measures, which is seen as the most crucial point for contactless fingerprint sample quality assessment. 

In contrast, the 10 most important features of NFIQ 2.2 share only approx. 33\% feature importance. 
The most important feature \textit{Frequency Domain Analysis Standard Deviation} represents a one-dimensional signature of the ridge-valley structure. 
From Table \ref{tab:feature_importance} we can also see that the NFIQ 2.2 model puts high importance on minutiae count and quality (\eg \textit{FingerJet FX OSE COM Minutiae Count} and \textit{FingerJet FX OSE OCL Minutiae Quality}) as quality features. \rev{A possible cause for this is that NFIQ 2 was trained on a contact-based fingerprint database which contains a large portion of partial fingerprints, which include fewer minutiae. }

\rev{From the obtained importance map we can conclude, that the most important features in MCLFIQ merely address the fidelity of a fingerprint sample, \cf Section \ref{items:utilityetc}, whereas most important features of NFIQ 2.2 include both, character and fidelity. This is plausibel, because the contactless capturing process poses significantly more challenges than the contact-based one, which directly addresses fidelity. In other words it can be summarized that fidelity is a greater challenge for contactless fingerprints compared to character in most cases.}


Furthermore, we compare the sizes of both models. MCLFIQ is only 295.6KB, whereas the NFIQ 2.2 model has a size of 52.9MB. This is mainly caused by the unbalanced feature importance. Many trees in the random forest are very shallow, which leads to a smaller model. It should be noted that also the time required to load the model is positively affected, which is especially beneficial for mobile and embedded devices. 

\begin{table*}[]
\caption{EERs obtained on the considered subsets using two different recognition workflows. 
Note that the labels of the test datasets are introduced in Table \ref{tab:dbs}.}
\centering
\scriptsize

\begin{tabular}{cL{0.03\linewidth}L{0.03\linewidth}L{0.03\linewidth}L{0.03\linewidth}L{0.03\linewidth}L{0.03\linewidth}L{0.03\linewidth}|L{0.03\linewidth}L{0.03\linewidth}L{0.03\linewidth}L{0.03\linewidth}L{0.03\linewidth}L{0.03\linewidth}}
\toprule
& \textbf{AIT} &\multicolumn{2}{c}{\textbf{HDA}}&\multicolumn{4}{c|}{\textbf{ISPFD}}&\textbf{AIT} & \multicolumn{2}{c}{\textbf{FVC2006}} & \textbf{HDA} & \textbf{ISPFD}\\
           & \textbf{mobile} & \textbf{const} & \textbf{uncon} & \textbf{WI} & \textbf{NI} & \textbf{NO} & \textbf{WO} & \textbf{cb} & \textbf{DB2} & \textbf{DB3} & \textbf{cb} & \textbf{LS} \\
           \midrule
IDKit & 3.49       & 16.06     & 18.69     & 8.39      & 14.63     & 4.47      & 2.49      & 1.49    & 0.06         & 0.94         & 1.49            & 0.4       \\
Open-source  & 5.11       & 32.59     & 30.41     & 16.99     & 16.75     & 5.06      & 2.67      & 6.03    & 0.12         & 2.68         & 6.03            & 0.73      \\
NBIS       & 13.84      & 17.77     & 24.24     & 42.47     & 16.69     & 8.63      & 6.07      & 9.55    & 2.03         & 7.97         & 9.55            & 1.26      \\
\bottomrule
\end{tabular}
\label{tab:det}
\end{table*}

\begin{figure*}[!t]
	\centering
	\subfloat[AIT mobile]{\includegraphics[width=0.2475\linewidth]{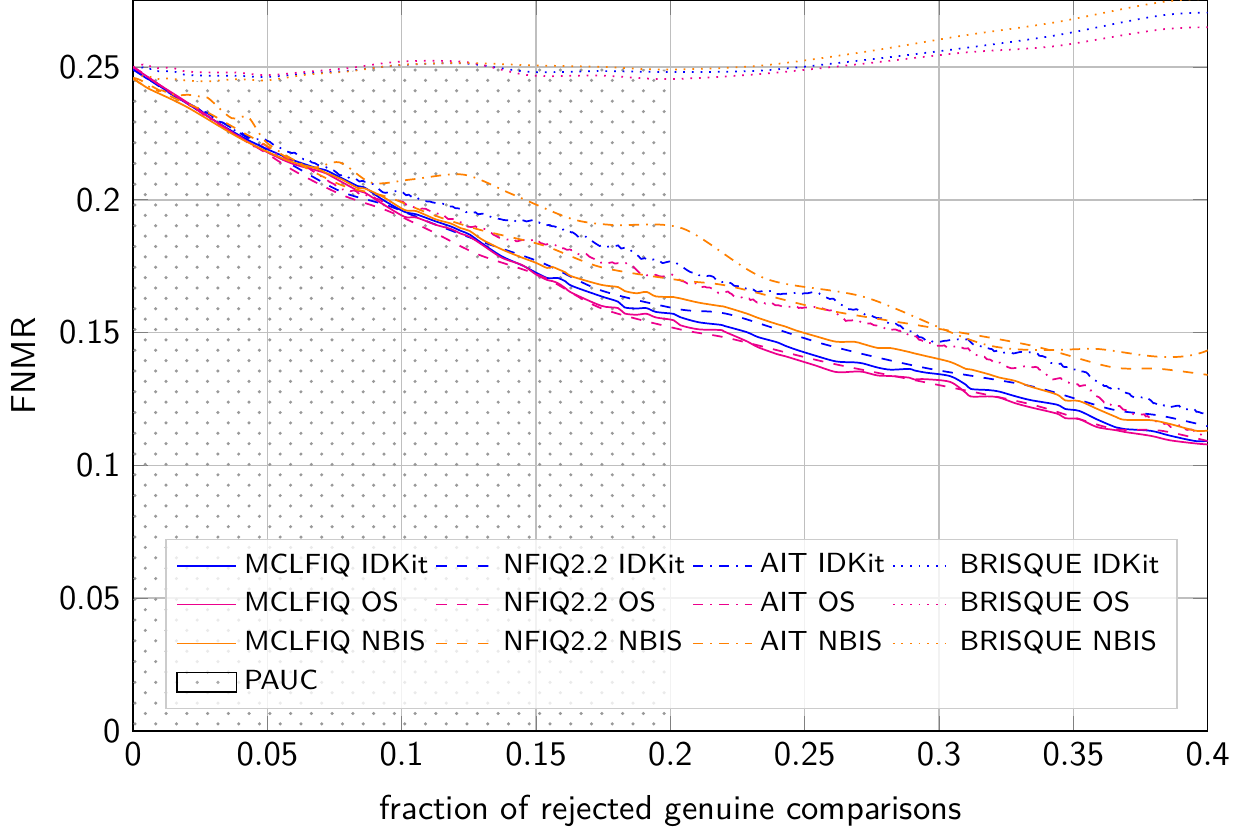}
	\label{fig:AIT_MOBILE_EDC}}
	\hfil
	\subfloat[HDA constrained]{\includegraphics[width=0.2475\linewidth]{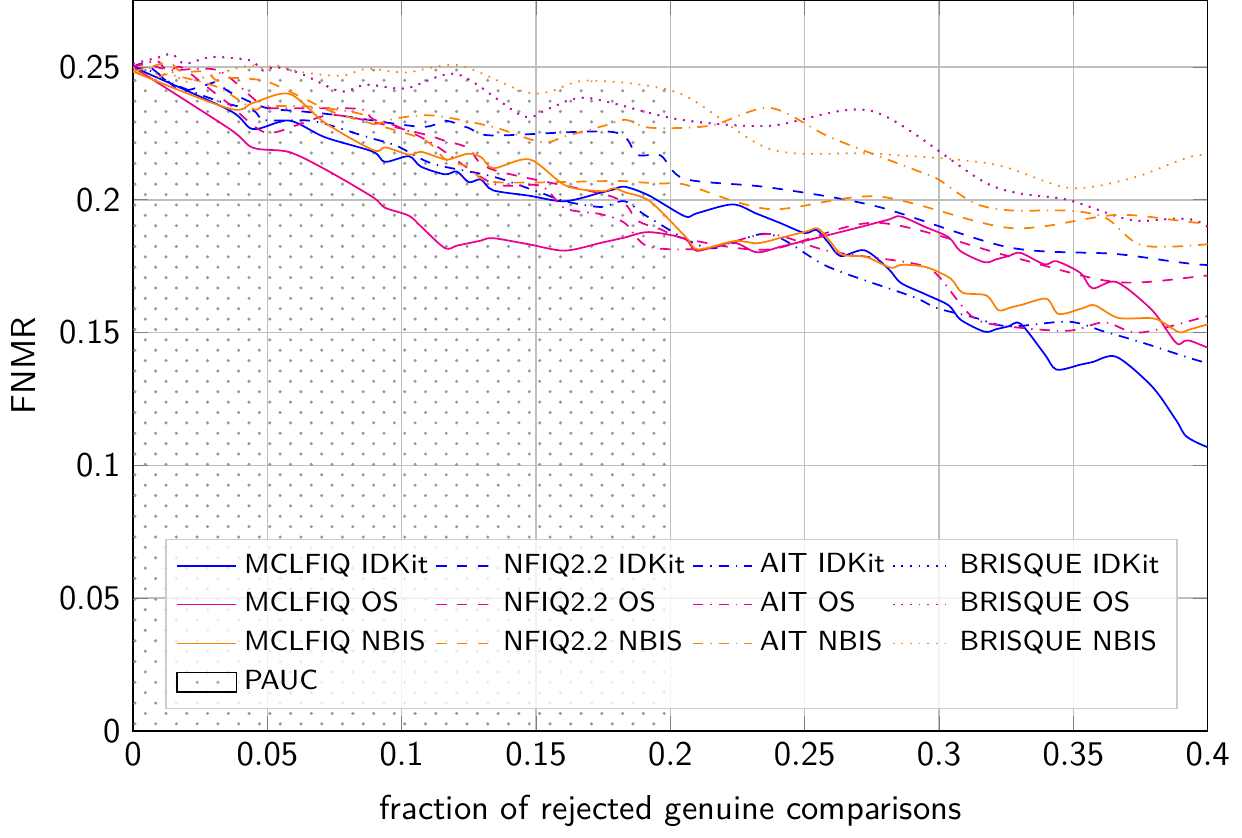}%
	\label{fig:HDA_CONST_EDC}}
	\hfil
	\subfloat[HDA unconstrained]{\includegraphics[width=0.2475\linewidth]{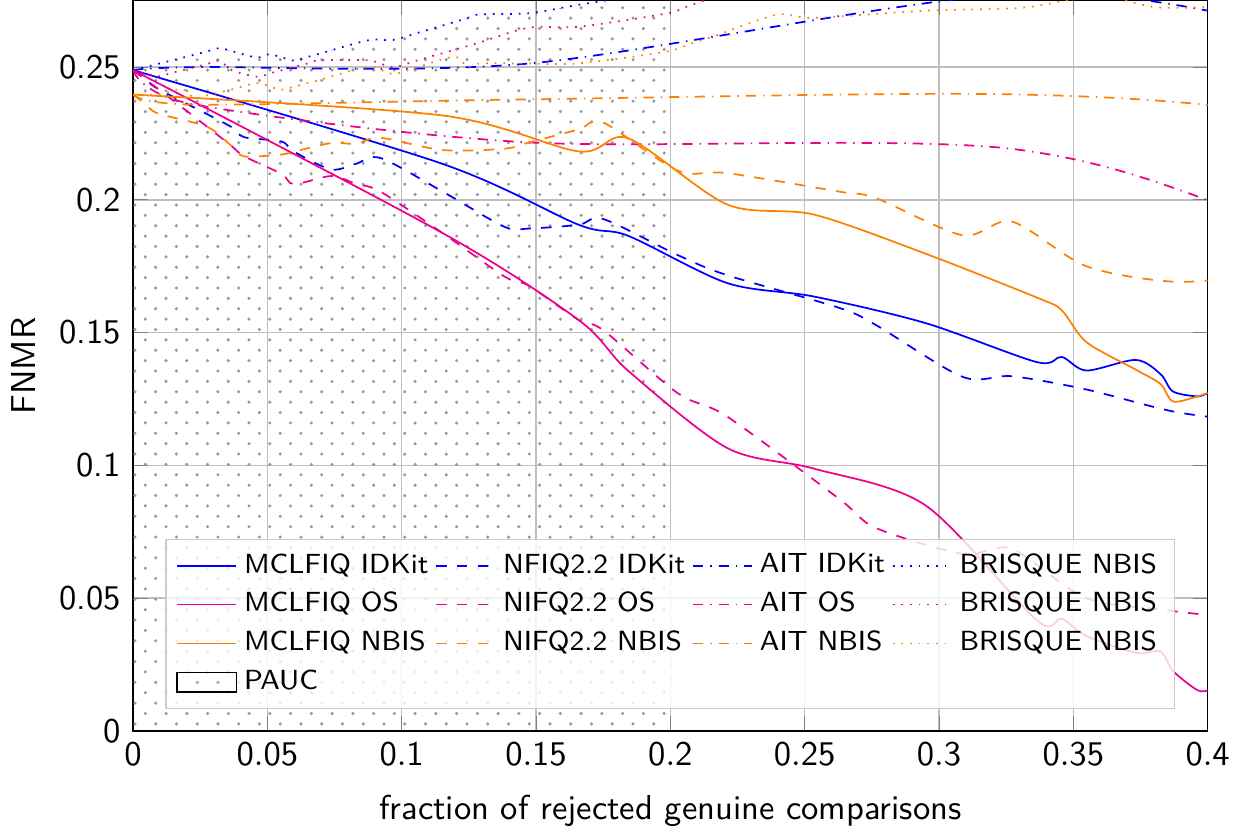}%
	\label{fig:HDA_UNCONST_EDC}}
	\hfil
	\subfloat[ISPFD NI]{\includegraphics[width=0.2475\linewidth]{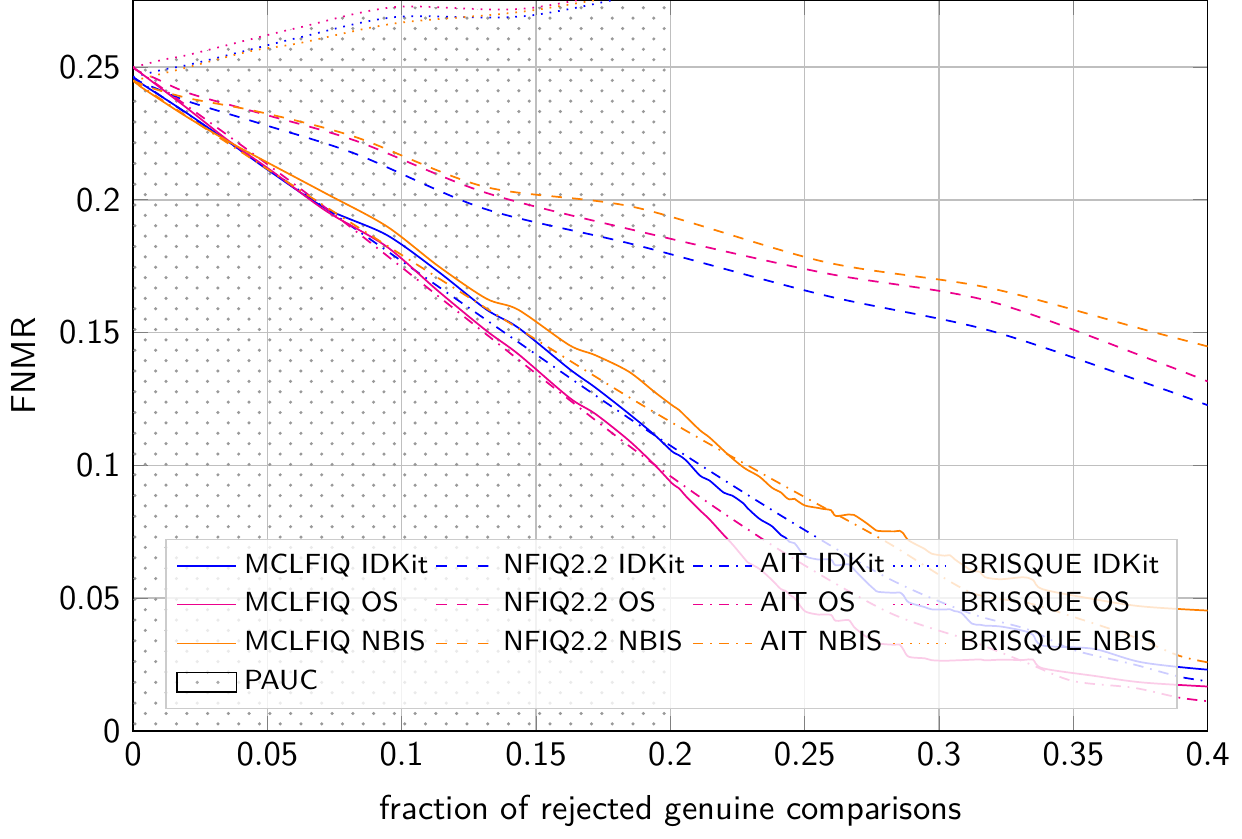}%
	\label{fig:ISPFD_NI_EDC}}
	\hfil
	\subfloat[ISPFD NO]{\includegraphics[width=0.2475\linewidth]{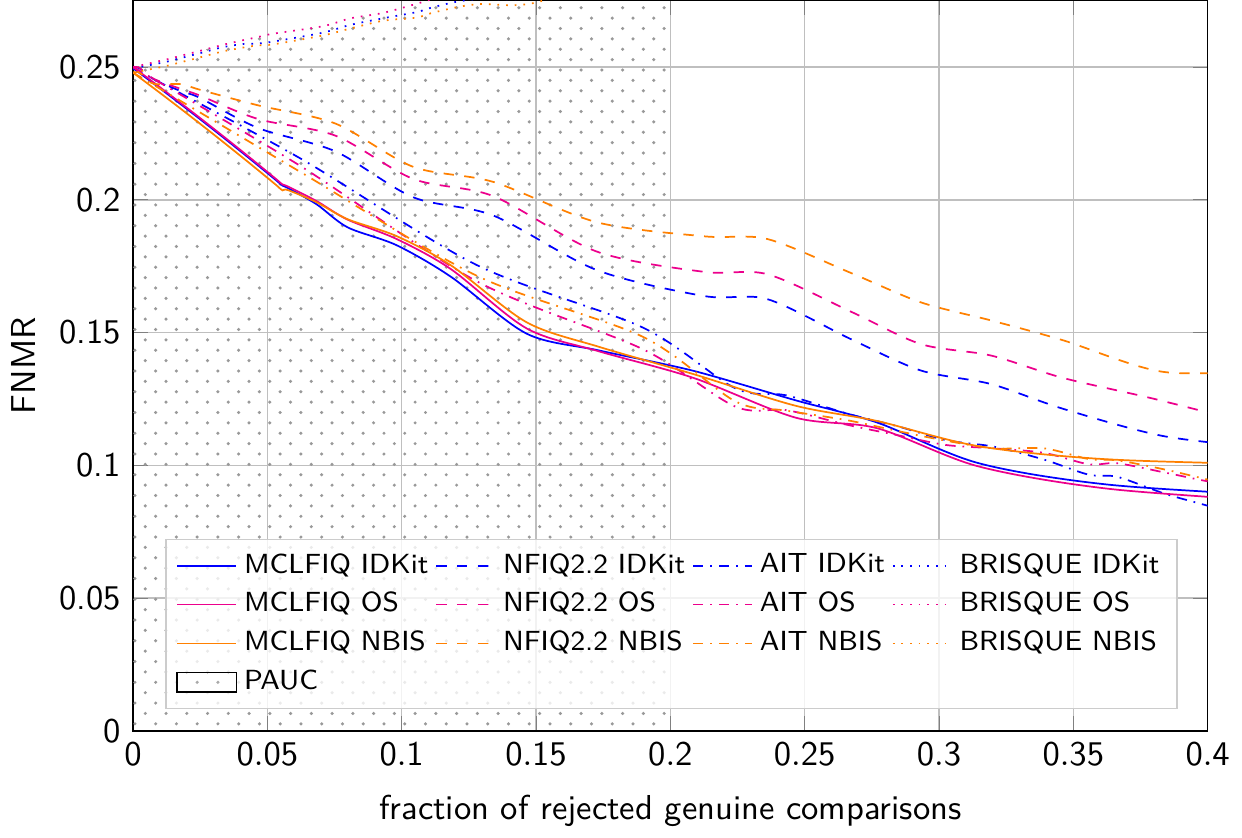}%
	\label{fig:ISPFD_NO_EDC}}
	\hfil
	\subfloat[ISPFD WI]{\includegraphics[width=0.245\linewidth]{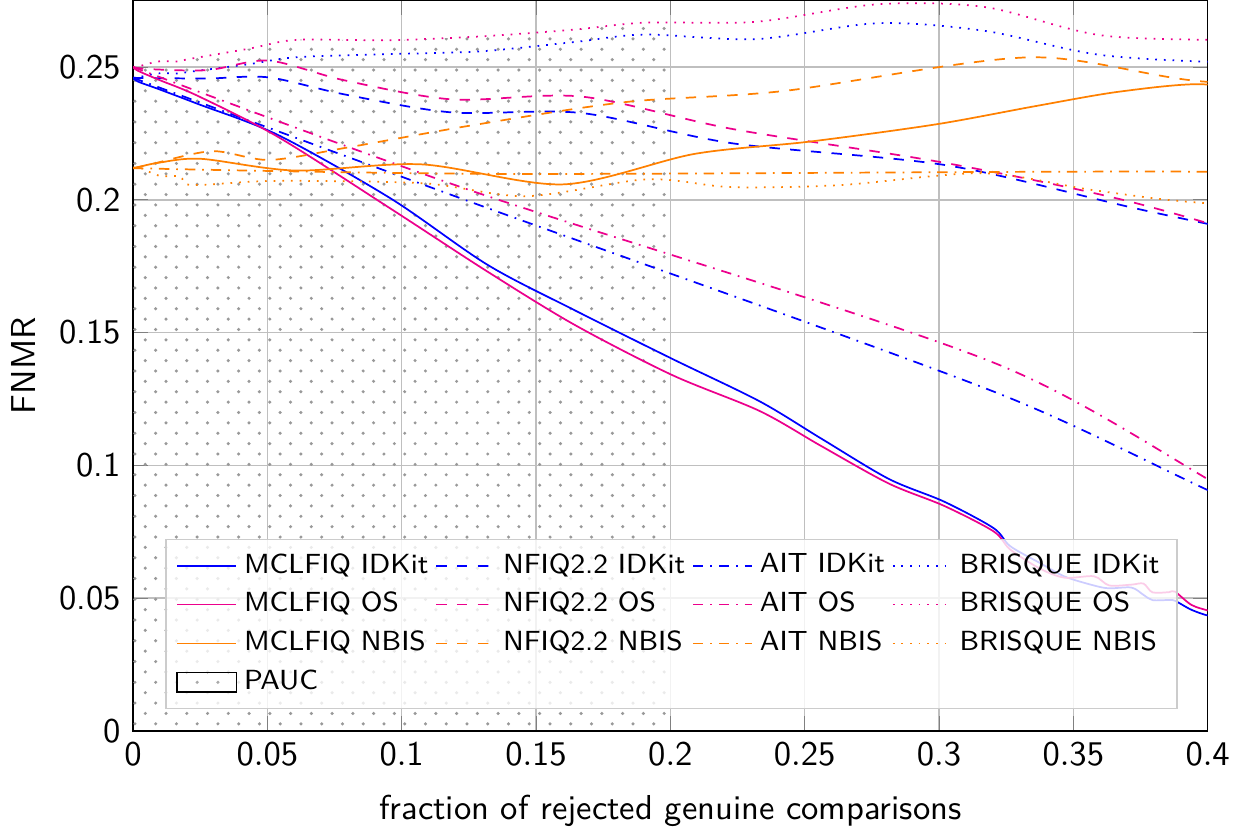}%
	\label{fig:ISPFD_WI_EDC}}
	\hfil
	\subfloat[ISPFD WO]{\includegraphics[width=0.2475\linewidth]{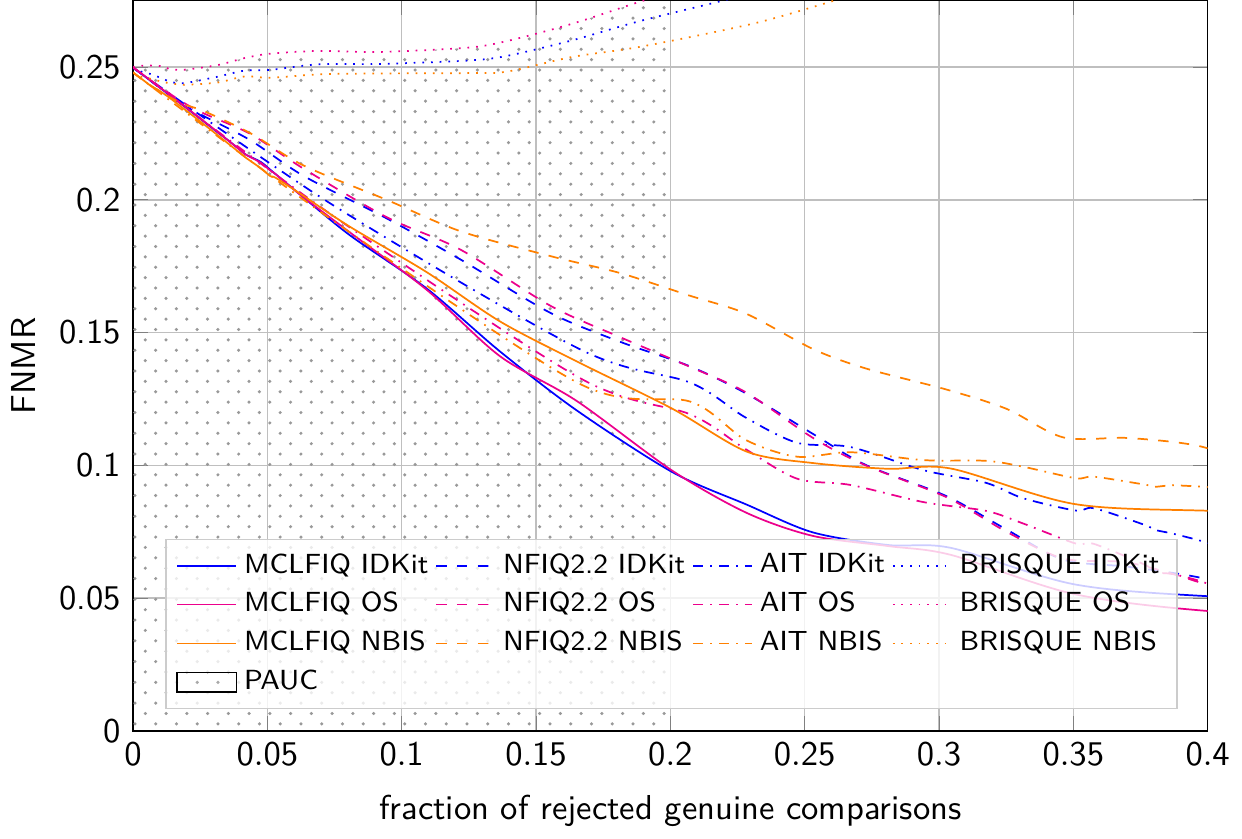}%
	\label{fig:ISPFD_WO_EDC}}
    \hfil 
	\caption{EDC curves obtained from the considered databases using the three quality assessment algorithms and the two recognition workflows. The EDC PAUC denotes the area which is considered during the EDC PAUC calculation. (OS: open-source recognition workflow, AIT: AIT sharpness metric)}
	\label{fig:edc}
\end{figure*}
\begin{table*}[t]
\caption{EDC PAUC the range [0, 0.2] obtained from the contactless databases using IDKit.}
\centering
 \scriptsize

\begin{tabular}{cL{0.05\linewidth}L{0.05\linewidth}L{0.05\linewidth}L{0.05\linewidth}L{0.05\linewidth}L{0.05\linewidth}L{0.05\linewidth}L{0.05\linewidth}L{0.05\linewidth}}
\toprule
&\multicolumn{4}{c}{\textbf{ISPFD}}&\multicolumn{2}{c}{\textbf{HDA}}&\textbf{AIT}&\multirow{2}{*}{\textbf{avg}}&\multirow{2}{*}{\textbf{std} \textbf{dev}}\\
       & \textbf{WI} & \textbf{NI} & \textbf{NO} & \textbf{WO} & \textbf{const} & \textbf{uncon} & \textbf{mobile} &   &  \\
\midrule
MCLFIQ        & 0.3913    & 0.3582    & 0.3662    & 0.3450    & 0.4358    & 0.4347    & 0.3973     & 0.3898 & 0.0360  \\
NFIQ 2.2      & 0.4754    & 0.4210    & 0.4120    & 0.3804    & 0.4623    & 0.4183    & 0.3945     & 0.4234 & 0.0343  \\
AIT sharpness & 0.4175    & 0.3534    & 0.3916    & 0.3681    & 0.4374    & 0.5025    & 0.4107     & 0.4116 & 0.0494  \\
BRISQUE       & 0.5098    & 0.5293    & 0.5375    & 0.5068    & 0.4919    & 0.5263    & 0.4972     & 0.5141 & 0.0172 \\
\bottomrule
\end{tabular}
	\label{tab:edc_auc_idkit}

\end{table*}

\subsection{Biometric Performance}
\rev{
First, we discuss the biometric performance for each database in combination with both recognition workflows. From Figure \ref{fig:det} and Table \ref{tab:det} we can observe two general trends: First, we see that the contact-based databases in general have a lower EER compared to the contactless ones, which is expected. Second, the considered identification workflows have different performance on the tested databases. The IDKit algorithm works rather robust on all the databases, whereas the open source workflows fall behind. Here, FingerNet + SourceAFIS are still more accurate than NBIS.} 

\rev{The DET plots in Figure \ref{fig:det} show the challenging characteristics of all the considered databases. Except the ISPFD WO, LS sub-database and the FVC2006 DB2 which have a good performance, all databases show a fair performance. Especially, the databases which were captured in an indoor environment (ISPFD WI and both HDA databases) have a poor performance. It should also be noted that the COTS system achieves lower EERs compared to the open-source workflow, especially on challenging data. 
This challenging characteristic of the  databases is highly suited for our experiments because the predictive power of the EDC method can be evaluated best on databases of heterogeneous quality. This means high-quality gains can be achieved by discarding samples of low quality.
}

\subsection{Predictive Power}

\begin{table*}[t]
\caption{EDC PAUC in range [0, 0.2] obtained from the contactless databases using the open-source method.}
	\centering
 \scriptsize

\begin{tabular}{cL{0.05\linewidth}L{0.05\linewidth}L{0.05\linewidth}L{0.05\linewidth}L{0.05\linewidth}L{0.05\linewidth}L{0.05\linewidth}L{0.05\linewidth}L{0.05\linewidth}}
\toprule
&\multicolumn{4}{c}{\textbf{ISPFD}}&\multicolumn{2}{c}{\textbf{HDA}}&\textbf{AIT}&\multirow{2}{*}{\textbf{avg}}&\multirow{2}{*}{\textbf{std} \textbf{dev}}\\
       & \textbf{WI} & \textbf{NI} & \textbf{NO} & \textbf{WO} & \textbf{const} & \textbf{uncon} & \textbf{mobile} &   &  \\\midrule
MCLFIQ        & 0.3867    & 0.3490    & 0.3684    & 0.3463    & 0.4065    & 0.3869    & 0.3908     & 0.3764 & 0.0225  \\
NFIQ 2.2      & 0.4857    & 0.4307    & 0.4227    & 0.3847    & 0.4442    & 0.3830    & 0.3886     & 0.4199 & 0.0379  \\
AIT sharpness & 0.4263    & 0.3474    & 0.3823    & 0.3559    & 0.4430    & 0.4508    & 0.4043     & 0.4014 & 0.0411  \\
BRISQUE       & 0.5204    & 0.5360    & 0.5422    & 0.5175    & 0.4869    & 0.5123    & 0.4974     & 0.5161 & 0.0196  \\
	\bottomrule
\end{tabular}
\label{tab:edc_auc_fingernet}
\end{table*}

\begin{table*}[t]
\caption{EDC PAUC in range [0, 0.2] obtained from the contactless databases using the NBIS algorithms. }
	\centering
 \scriptsize

\begin{tabular}{cL{0.05\linewidth}L{0.05\linewidth}L{0.05\linewidth}L{0.05\linewidth}L{0.05\linewidth}L{0.05\linewidth}L{0.05\linewidth}L{0.05\linewidth}L{0.05\linewidth}}
	\toprule
&\multicolumn{4}{c}{\textbf{ISPFD}}&\multicolumn{2}{c}{\textbf{HDA}}&\textbf{AIT}&\multirow{2}{*}{\textbf{avg}}&\multirow{2}{*}{\textbf{std} \textbf{dev}}\\
       & \textbf{WI} & \textbf{NI} & \textbf{NO} & \textbf{WO} & \textbf{const} & \textbf{uncon} & \textbf{mobile} &   &  \\
       \midrule
MCLFIQ        & 0.4228    & 0.3684    & 0.3690    & 0.3600    & 0.4419    & 0.4642    & 0.3979     & 0.4035 & 0.0406  \\
NFIQ 2.2       & 0.4491    & 0.4359    & 0.4341    & 0.4017    & 0.4504    & 0.4456    & 0.4039     & 0.4315 & 0.0206  \\
AIT sharpness & 0.4206    & 0.3589    & 0.3836    & 0.3520    & 0.4656    & 0.4726    & 0.4173     & 0.4101 & 0.0480  \\
BRISQUE       & 0.4130    & 0.5286    & 0.5357    & 0.4978    & 0.4932    & 0.4955    & 0.4972     & 0.4944 & 0.0398 \\
\bottomrule
\end{tabular}
\label{tab:edc_auc_nbis}
\end{table*}

\begin{figure*}[!t]
	\centering
	\subfloat[AIT contact-based]{\includegraphics[width=0.2475\linewidth]{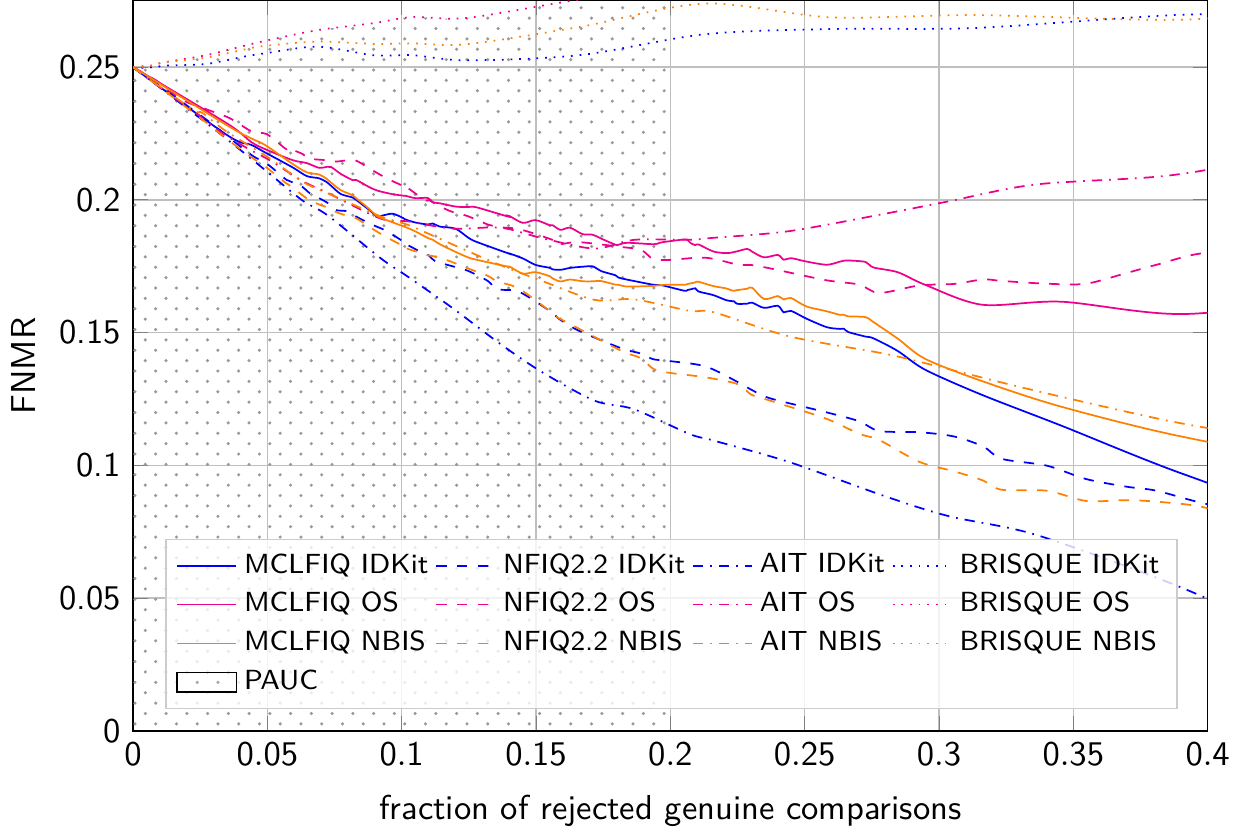}
	\label{fig:AIT_MOBILE_EDC}}
	\hfil
	\subfloat[HDA cb]{\includegraphics[width=0.2475\linewidth]{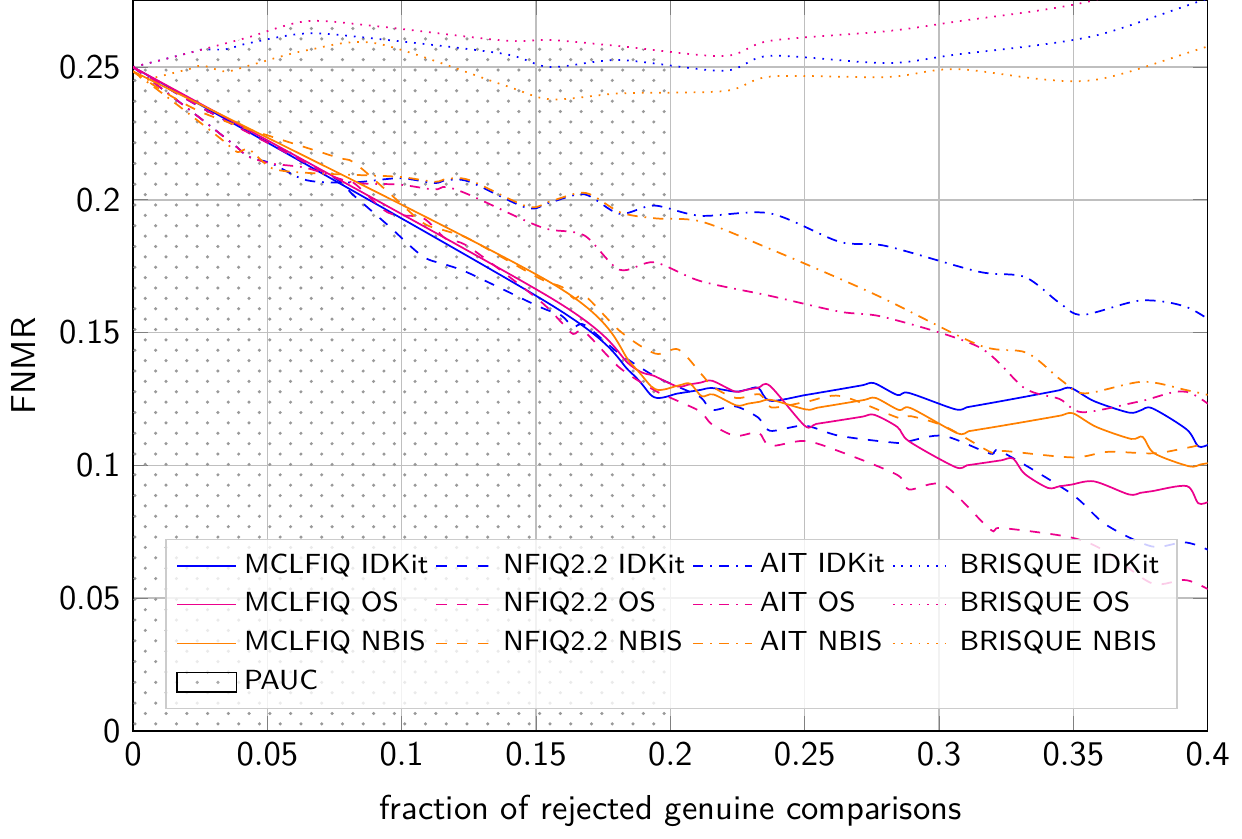}%
	\label{fig:HDA_CONST_EDC}}
	\hfil
	\subfloat[ISPFD LS]{\includegraphics[width=0.2475\linewidth]{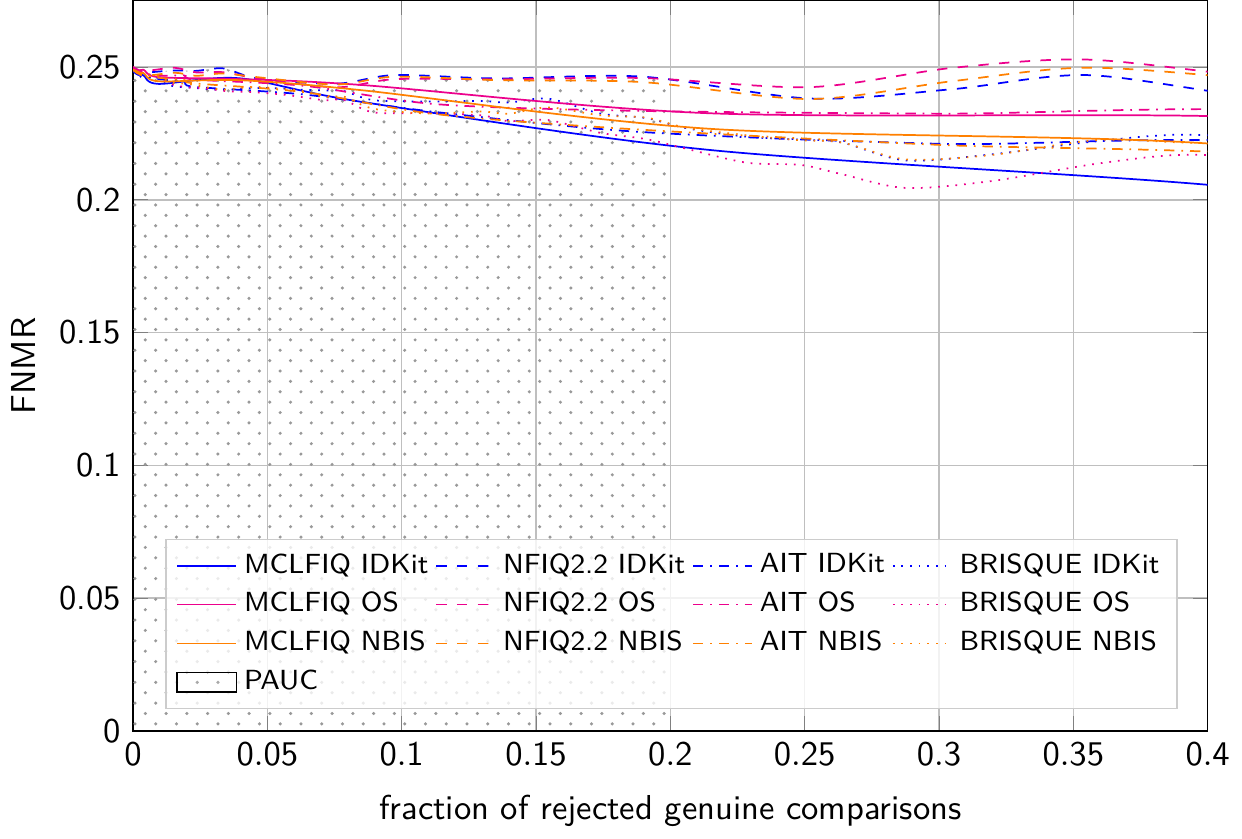}%
	\label{fig:HDA_UNCONST_EDC}}
	
	\subfloat[FVC 2006 DB2]{\includegraphics[width=0.2475\linewidth]{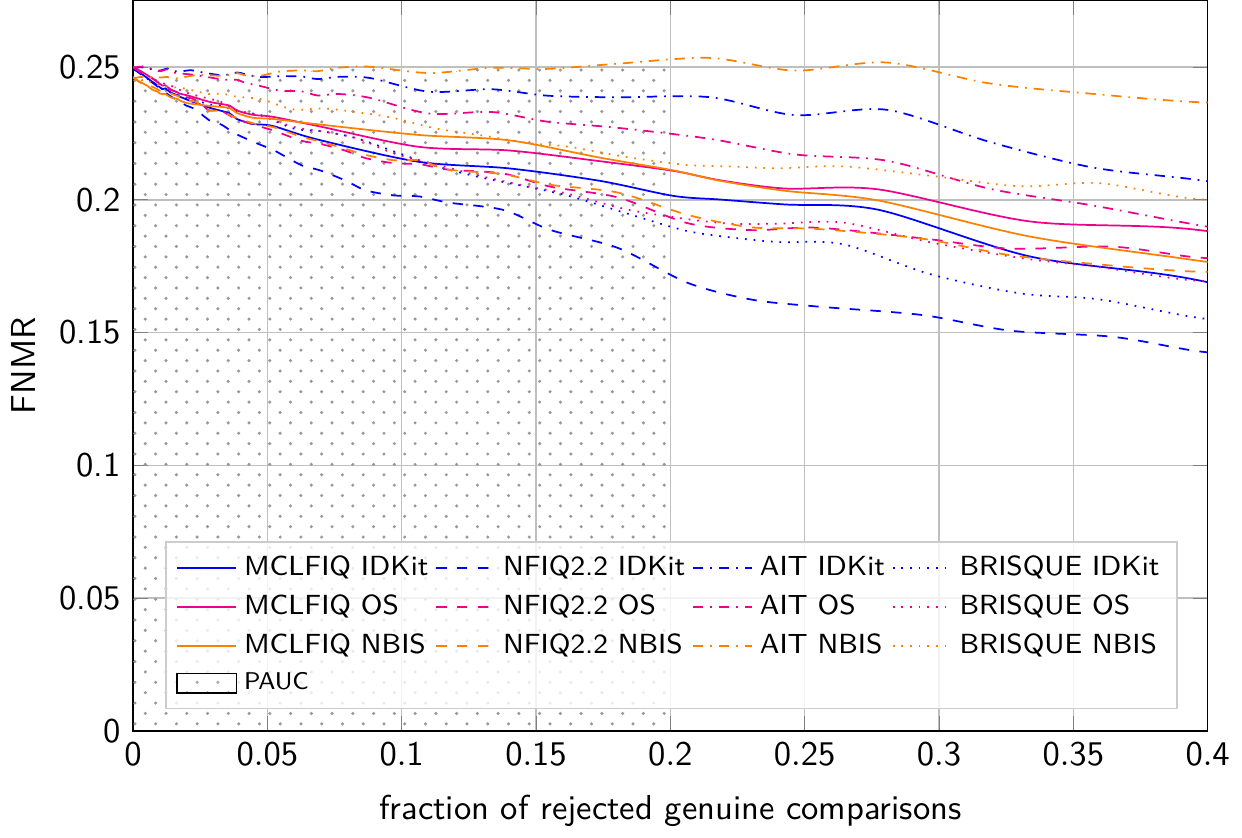}%
	\label{fig:ISPFD_NI_EDC}}
	\hfil
	\subfloat[FVC 2006 DB3]{\includegraphics[width=0.2475\linewidth]{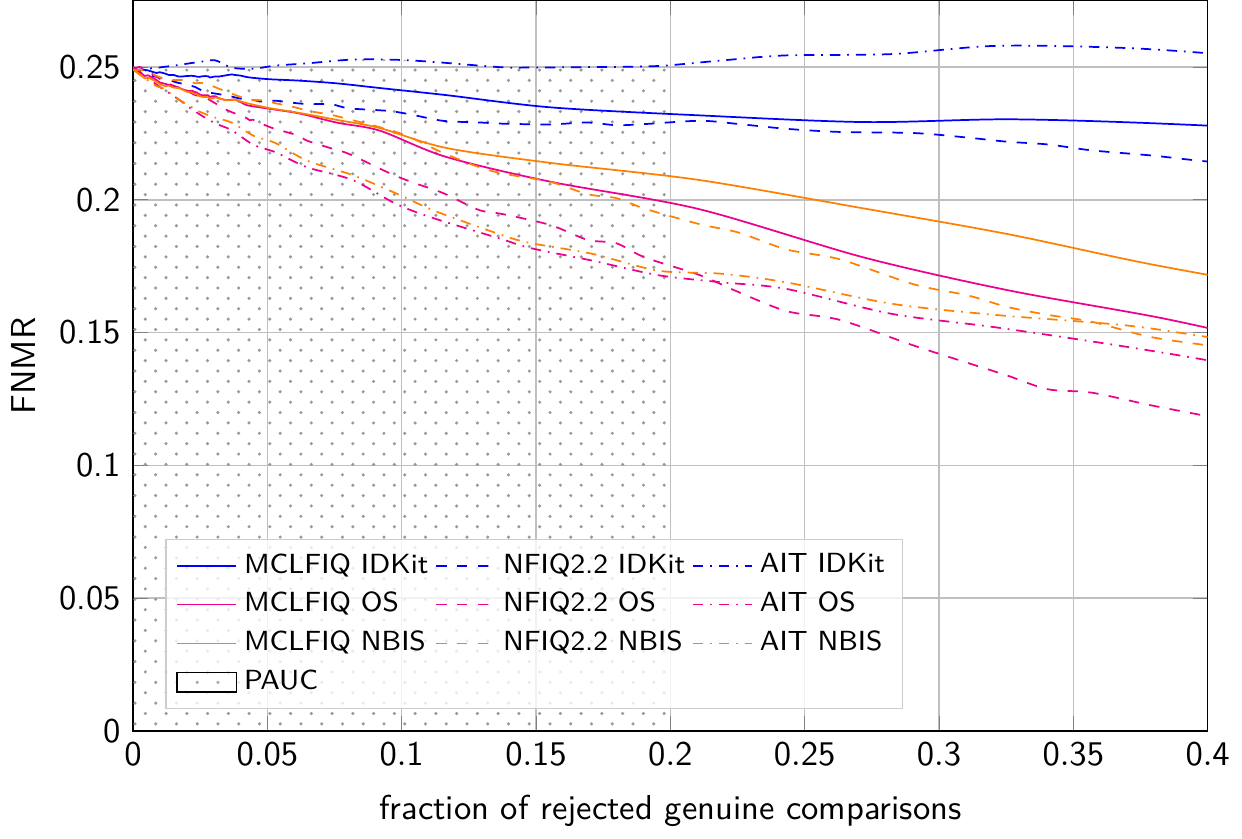}%
	\label{fig:ISPFD_NO_EDC}}
	\caption{EDC curves obtained from the considered databases using the three quality assessment algorithms and the two recognition workflows. The EDC PAUC denotes the area which is considered during the EDC PAUC calculation. It should be noted that FVC 2006 DB3 it is not possible to compute reasonable BRISQUE scores, which is why the curves are missing. (OS: open-source recognition workflow, AIT: AIT sharpness metric)}
	\label{fig:edc_cd}
\end{figure*}

\begin{table}[t]
\caption{ CB EDC PAUC in range [0, 0.2] obtained from the contact-based  databases using IDKit.}
\scriptsize

\begin{tabular}{L{0.1635\linewidth}L{0.0657\linewidth}L{0.0657\linewidth}L{0.0657\linewidth}L{0.0657\linewidth}L{0.0657\linewidth}L{0.0657\linewidth}L{0.0657\linewidth}}
\toprule
&\multicolumn{2}{c}{\textbf{FVC2006}}&\textbf{ISPFD}&\textbf{AIT}&\textbf{HDA}&\multirow{2}{*}{\textbf{avg}}&\textbf{std} \\
       & \textbf{DB2}& \textbf{DB3} & \textbf{LS}  & \textbf{cb} & \textbf{cb} &   & \textbf{dev} \\
\midrule
MCLFIQ        & 0.4387       & 0.4813       & 0.4681      & 0.3964  & 0.3914         & 0.4352 & 0.0408  \\
NFIQ 2.2       & 0.4112       & 0.4678       & 0.4928      & 0.3751  & 0.3798         & 0.4254 & 0.0528  \\
AIT sharpness & 0.4871       & 0.5017       & 0.4694      & 0.3506  & 0.4206         & 0.4459 & 0.0614  \\
BRISQUE       & 0.4347       & \hspace{1.5mm}--           & 0.4768      & 0.5085  & 0.5126         & 0.4831 & 0.0360 \\
\bottomrule
\end{tabular}
\label{tab:edc_auc_idkit_cb}

\end{table}

We evaluate the predictive power of each quality assessment algorithm in terms of EDC curves, as introduced in Section \ref{sec:sample_quality}. For the EDC computations, it is required to set an initial FNMR. Here, a good practice is to consider approximately the EER as initial FNMR. For this reason, we set the initial FNMR to 0.25\% for all experiments. For better comparability, we also report the EDC PAUC, which refers to the area under the curve in the range between $[0, 0.2]$. 

The EDC curves (\cf Figure \ref{fig:edc}) show that all considered quality assessment algorithms show reasonable results \rev{on some of the tested databases.} 
\rev{
However, the results indicate that the retrained BRISQUE performs poor on all contactless databases except HDA constrained.  Also, the HDA unconstrained sub-database in combination with AIT sharpness and IDKit indicates poor performance. 
All other EDC curves decrease from the starting point, which indicates a lower FNMR by discarding samples which were identified as low quality by the quality assessment method. Also, there is no huge difference between the EDCs obtained by using the COTS algorithm and both  open-source workflow. From this, we can summarize that the predictive power of the considered quality assessment algorithm is independent of the used recognition workflow. 
}

In more detail, the EDC curves also show that MCLFIQ performs best if the average of every EDC PAUC is considered, \cf Tables \ref{tab:edc_auc_idkit} -- \ref{tab:edc_auc_nbis}. Especially on the ISPFD database, NFIQ 2.2 has an inferior performance compared to both other methods. The AIT sharpness metric performs slightly better on the ISPFD NI sub-database, but worse on all other databases. In summary, MCLFIQ has the best overall performance on the ISPFD database. 

\begin{table}[t]
\caption{EDC PAUC in range [0, 0.2] obtained from the contact-based  databases using the open-source method. }
\scriptsize
\begin{tabular}{L{0.1635\linewidth}L{0.0657\linewidth}L{0.0657\linewidth}L{0.0657\linewidth}L{0.0657\linewidth}L{0.0657\linewidth}L{0.0657\linewidth}L{0.0657\linewidth}}
\toprule
&\multicolumn{2}{c}{\textbf{FVC2006}}&\textbf{ISPFD}&\textbf{AIT}&\textbf{HDA}&\multirow{2}{*}{\textbf{avg}}& \textbf{std} \\
       & \textbf{DB2}& \textbf{DB3} & \textbf{LS}  & \textbf{cb} & \textbf{cb} &   & \textbf{dev} \\
\midrule
MCLFIQ        & 0.4493       & 0.4437       & 0.4828      & 0.4142  & 0.3956         & 0.4371 & 0.0336  \\
NFIQ 2.2       & 0.4341       & 0.4202       & 0.4916      & 0.4137  & 0.3843         & 0.4288 & 0.0396 \\
AIT sharpness & 0.4732       & 0.4036       & 0.4776      & 0.4032  & 0.4111         & 0.4337 & 0.0382  \\
BRISQUE       & 0.4363       & \hspace{1.5mm}--        & 0.4683      & 0.5324  & 0.5212         & 0.4896 & 0.0452  \\
\bottomrule
\end{tabular}
\label{tab:edc_auc_os_cb}
\end{table}

\begin{figure*}[!t]
	\centering
	\includegraphics[width=0.75\linewidth]{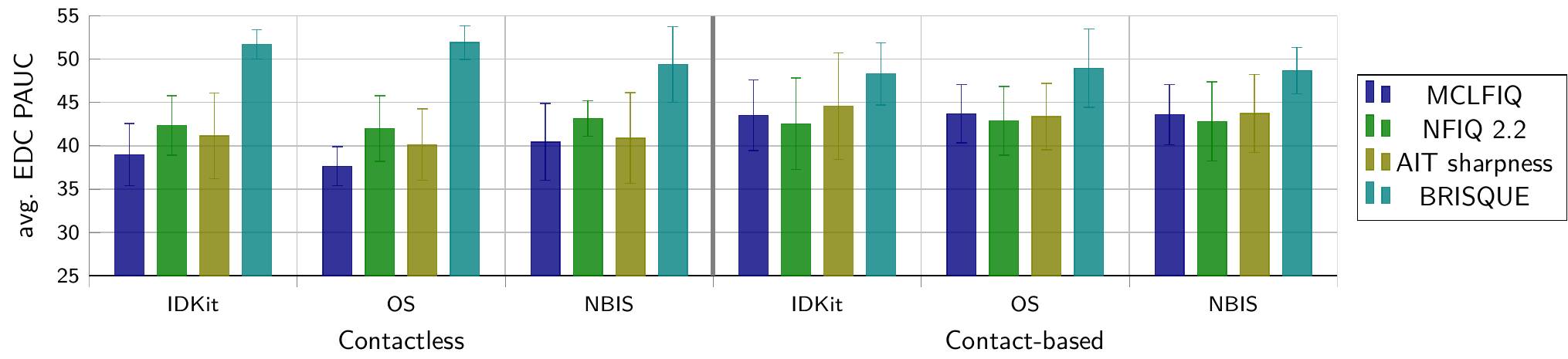}%
	\caption{Overview of the average EDC PAUC incl. standard deviation obtained using the different quality assessment algorithms and recognition workflows.}
	\label{fig:edc_auc_bar}
\end{figure*}
Considering the HDA database, it is observable that the EDC curves are not decreasing as monotonously as the others. Also, on this database, the recognition workflow seems to have a major impact on the predictive power. These findings can be attributed to the small total number of samples in the database with the highly challenging characteristic of the database. On the constrained subset, we can see that NFIQ 2.2 performs worst, whereas the AIT sharpness and MCLFIQ are very close. Most notably, the predictive performance of the open-source workflow together with MCLFIQ performs better than all other combinations. 
On the unconstrained sub-database, the predictive power of every assessment algorithm is better in combination with the open-source workflow than with IDKit. Here, MCLFIQ and NFIQ 2.2 have a comparable EDC PAUC, whereas the AIT sharpness algorithm is worse.  

The EDCs obtained by the AIT mobile database are very close together. It can be seen that the AIT sharpness metric performs slightly worse compared to MCLFIQ and NFIQ 2.2. Again, all quality assessment algorithms seem to have a better predictive power in combination with the open-source workflow.  

\rev{To support our proposal we also conduct a counter experiment by benchmarking the considered quality assessment algorithms on contact-based databases. Figure \ref{fig:edc_cd} presents the results in terms of EDC curves. For MCLFIQ and NFIQ 2.2, the obtained results show the expected result: NFIQ 2.2 shows in general a lower EDC PAUC  than MCLFIQ, \cf Tables \ref{tab:edc_auc_idkit_cb} -- \ref{tab:edc_auc_nbis_cb}. Most notably, both the AIT sharpness metric and the BRISQUE algorthithm show good results in some experiments. Most likely this is caused by special database properties seen in the FVC 2006 DB3 and should be further investigated. }

Figure \ref{fig:edc_auc_bar} presents a visual representation of the average EDC PAUC, including the standard deviations obtained from the experiments. From the charts, we can see that the predictive power achieved by MCLFIQ is on average much lower compared to NFIQ 2.2 and the AIT sharpness metric\rev{, whereas BRISQE shows degraded performance.} Also, the standard deviation is much lower compared to the AIT sharpness metric. NFIQ 2.2 combined with IDKit has a slightly lower standard deviation at a worse predictive power. From the results, we can conclude, that MCLFIQ works more accurate and more robust compared to both other methods. 

\begin{table}[t]
\caption{EDC PAUC in range [0, 0.2] obtained from the contact-based databases using the NBIS algorithms.}
\scriptsize
\begin{tabular}{L{0.1635\linewidth}L{0.0657\linewidth}L{0.0657\linewidth}L{0.0657\linewidth}L{0.0657\linewidth}L{0.0657\linewidth}L{0.0657\linewidth}L{0.0657\linewidth}}
\toprule
&\multicolumn{2}{c}{\textbf{FVC2006}}&\textbf{ISPFD}&\textbf{AIT}&\textbf{HDA}&\multirow{2}{*}{\textbf{avg}}&\textbf{std}\\
       & \textbf{DB2}& \textbf{DB3} & \textbf{LS}  & \textbf{cb} & \textbf{cb} &   & \textbf{dev} \\
\midrule
MCLFIQ        & 0.4517       & 0.4503       & 0.4774      & 0.3938  & 0.4067         & 0.4360 & 0.0346  \\
NFIQ2.2       & 0.4351       & 0.4452       & 0.4911      & 0.3738  & 0.3946         & 0.4280 & 0.0458  \\
AIT sharpness & 0.4979       & 0.4077       & 0.4703      & 0.3895  & 0.4209         & 0.4373 & 0.0452  \\
BRISQUE       & 0.4585       & \hspace{1.5mm}--           & 0.4735      & 0.5195  & 0.4967         & 0.4870 & 0.0267 \\
\bottomrule
\end{tabular}
\label{tab:edc_auc_nbis_cb}
\end{table}
\section{Conclusion and Future Work}
\label{sec:conclusion}
Contactless fingerprint recognition has gained a lot of attention in recent years. However, quality assessment of contactless fingerprints remains a not yet sufficiently covered research topic.
With MCLFIQ we propose a quality assessment algorithm for mobile contactless fingerprint samples. Based on the NFIQ 2 framework, a new random forest model is trained on synthetic contactless fingerprint samples which were generated using the SynCoLFinGer method. For testing the proposed model, four mobile contactless fingerprint databases are selected. Moreover, three different recognition workflows are employed for the experiments. The predictive power of MCLFIQ is benchmarked against NFIQ 2, a sharpness-based quality assessment algorithm and BRISQUE. 

Our experiments show that MCLFIQ is an effective method for predicting the quality in a mobile contactless fingerprint recognition workflow. MCLFIQ significantly outperforms NFIQ 2.2, the sharpness-based quality assessment and BRISQUE in terms of predictive performance and robustness. Also, it requires less memory than NFIQ 2.2. As stated, the MCLFIQ model is made publicly available to ensure the reproducibility of this work. 

\rev{Since the research area of contactless fingerprint recognition still lacks a standardized quality assessment algorithm, we suggest considering the proposed MCLFIQ method as a first baseline for research on a standardized quality assessment tool for contactless fingerprint samples.}

\rev{Further reseach should focus on the acquisition of a large contactless fingerprint database for training and testing quality assement methods. It is assumed that our proposal works even better when the re-training is done on real data.  
Furthermore, new research directions for contactless fingerprint quality assessment like CNN-based methods should be studied. }
\section*{Acknowledgments}
This research work has been funded by the German Federal Ministry of Education and Research and the Hessian Ministry of Higher Education, Research, Science and the Arts within their joint support of the National Research Center for Applied Cybersecurity ATHENE as well as the Austrian Research Promotion Agency (FFG) grant number 873462, project name “BioCapture”.



 
%

\bibliographystyle{IEEEtran}
\bibliography{cas-refs}

\newpage

{\appendices
\section*{Appendix A}
\begin{table}[H]
\caption{Feature Importance of MCLFIQ and NFIQ 2.2 }
	\label{tab:feature_importance_full}
	\centering
	\scriptsize
\begin{tabular}{cL{0.15\linewidth}L{0.15\linewidth}}
\hline
\textbf{Feature} & \textbf{MCLFIQ} & \textbf{NFIQ 2.2} \\\hline
Frequency Domain Analysis Bin 0 &2.13\%&0.89\%\\
Frequency Domain Analysis Bin 1 &0.04\%&0.85\%\\
Frequency Domain Analysis Bin 2 &0.04\%&0.82\%\\
Frequency Domain Analysis Bin 3 &0.04\%&0.85\%\\
Frequency Domain Analysis Bin 4 &0.04\%&0.83\%\\
Frequency Domain Analysis Bin 5 &0.04\%&0.88\%\\
Frequency Domain Analysis Bin 6 &0.04\%&1.20\%\\
Frequency Domain Analysis Bin 7 &0.08\%&1.49\%\\
Frequency Domain Analysis Bin 8 &0.46\%&1.50\%\\
Frequency Domain Analysis Bin 9 &0.04\%&2.28\%\\
Frequency Domain Analysis Mean &1.07\%&2.97\%\\
Frequency Domain Analysis Standard Deviation&0.17\%&6.72\%\\
FingerJet FX OSE COM Minutiae Count &0.05\%&4.40\%\\
FingerJet FX OSE Total Minutiae Count &0.45\%&2.75\%\\
FingerJet FX OSE Mu Minutiae Quality &0.04\%&1.54\%\\
FingerJet FX OSE OCL Minutiae Quality &5.46\%&3.96\%\\
ROI Area Mean &0.65\%&1.67\%\\
Local Clarity Score Bin 0 &0.04\%&0.00\%\\
Local Clarity Score Bin1 &0.77\%&1.00\%\\
Local Clarity Score Bin 2 &0.04\%&0.87\%\\
Local Clarity Score Bin 3 &0.04\%&0.90\%\\
Local Clarity Score Bin 4 &0.04\%&0.84\%\\
Local Clarity Score Bin 5 &0.05\%&0.91\%\\
Local Clarity Score Bin 6 &0.04\%&1.14\%\\
Local Clarity Score Bin 7 &0.05\%&2.42\%\\
Local Clarity Score Bin 8 &0.10\%&2.39\%\\
Local Clarity Score Bin 9 &0.05\%&1.11\%\\
Local Clarity Score Mean &1.01\%&1.74\%\\
Local Clarity Score Standard Deviation &0.04\%&1.37\%\\
MMB &0.25\%&1.56\%\\
MU &0.59\%&1.57\%\\
Orientation Certainty Level Bin 0 &7.60\%&0.97\%\\
Orientation Certainty Level Bin 1 &0.04\%&0.93\%\\
Orientation Certainty Level Bin 2 &0.05\%&0.89\%\\
Orientation Certainty Level Bin 3 &0.04\%&0.85\%\\
Orientation Certainty Level Bin 4 &0.04\%&0.90\%\\
Orientation Certainty Level Bin 5 &0.14\%&0.87\%\\
Orientation Certainty Level Bin 6 &1.29\%&0.92\%\\
Orientation Certainty Level Bin 7 &5.27\%&0.95\%\\
Orientation Certainty Level Bin 8 &1.56\%&1.26\%\\
Orientation Certainty Level Bin 9 &0.04\%&0.87\%\\
Orientation Certainty Level Mean &21.29\%&1.43\%\\
Orientation Certainty Level Standard Deviation &0.09\%&1.36\%\\
Orientation Flow Bin 0 &0.06\%&1.60\%\\
Orientation Flow Bin 1 &0.24\%&1.61\%\\
Orientation Flow Bin 2 &0.04\%&1.77\%\\
Orientation Flow Bin 3 &0.04\%&1.42\%\\
Orientation Flow Bin 4 &0.04\%&1.09\%\\
Orientation Flow Bin 5 &0.04\%&1.12\%\\
Orientation Flow Bin 6 &0.04\%&1.00\%\\
Orientation Flow Bin 7 &0.05\%&1.02\%\\
Orientation Flow Bin 8 &0.08\%&1.00\%\\
Orientation Flow Bin 9 &0.04\%&1.02\%\\
Orientation Flow Mean &0.12\%&1.49\%\\
Orientation Flow Standard Deviation &0.04\%&1.32\%\\
ROI Relative Orientation Map Coherence Sum &38.56\%&1.37\%\\
ROI Orientation Map Coherence Sum &7.45\%&1.82\%\\
Ridge Valley Uniformity Bin 0 &0.78\%&0.98\%\\
Ridge Valley Uniformity Bin 1 &0.15\%&1.07\%\\
Ridge Valley Uniformity Bin 2 &0.06\%&1.09\%\\
Ridge Valley Uniformity Bin 3 &0.04\%&1.68\%\\
Ridge Valley Uniformity Bin 4 &0.04\%&1.47\%\\
Ridge Valley Uniformity Bin 5 &0.06\%&1.70\%\\
Ridge Valley Uniformity Bin 6 &0.04\%&1.02\%\\
Ridge Valley Uniformity Bin 7 &0.32\%&0.99\%\\
Ridge Valley Uniformity Bin 8 &0.04\%&0.00\%\\
Ridge Valley Uniformity Bin 9 &0.04\%&0.00\%\\
Ridge Valley Uniformity Mean &0.04\%&3.32\%\\
Ridge Valley Uniformity Standard Deviation &0.07\%&2.43\%\\\hline
\end{tabular}
	
\end{table}

}

\vfill

\end{document}